\documentclass[sigconf]{acmart}

\AtBeginDocument{%
  \providecommand\BibTeX{{%
    \normalfont B\kern-0.5em{\scshape i\kern-0.25em b}\kern-0.8em\TeX}}}

\copyrightyear{2022}
\acmYear{2022}
\setcopyright{acmcopyright}\acmConference[ICPP '22]{51st International Conference
on Parallel Processing}{August 29-September 1, 2022}{Bordeaux, France}
\acmBooktitle{51st International Conference on Parallel Processing (ICPP '22), August
29-September 1, 2022, Bordeaux, France}
\acmPrice{15.00}
\acmDOI{10.1145/3545008.3545011}
\acmISBN{978-1-4503-9733-9/22/08}

\usepackage[ruled,vlined,linesnumbered]{algorithm2e}
\usepackage{multirow}
\usepackage{multicol}
\usepackage{subfigure}

\acmSubmissionID{pap106s1}

\begin{document}

%%
%% The "title" command has an optional parameter,
%% allowing the author to define a "short title" to be used in page headers.
\title{EmbRace: Accelerating Sparse Communication for Distributed Training of Deep Neural Networks}

%%
%% The "author" command and its associated commands are used to define
%% the authors and their affiliations.
%% Of note is the shared affiliation of the first two authors, and the
%% "authornote" and "authornotemark" commands
%% used to denote shared contribution to the research.
% \author{
% % 1st. author
%   Shengwei Li,
%   Zhiquan Lai,
%   Dongsheng Li,
%   Yiming Zhang,
%   Xiangyu Ye,
%   Yabo Duan
% }
% \affiliation{National Key Laboratory of Parallel and Distributed Processing\\
% Computer College, National University of Defense Technology
% \country{China}
% }
% \email{{lucasleesw9,sdiris}@gmail.com, {zqlai, dsli, xyye, yaboduan}@nudt.edu.cn}
\author{Shengwei Li}
\affiliation{
National Key Laboratory of Parallel and Distributed Processing\\
Computer College, National University of Defense Technology
\country{China}
}
\email{lucasleesw9@gmail.com}

\author{Zhiquan Lai}
\affiliation{
National Key Laboratory of Parallel and Distributed Processing\\
Computer College, National University of Defense Technology
\country{China}
}
\email{zqlai@nudt.edu.cn}

\author{Dongsheng Li}
\affiliation{
National Key Laboratory of Parallel and Distributed Processing\\
Computer College, National University of Defense Technology
\country{China}
}
\email{dsli@nudt.edu.cn}

\author{Yiming Zhang}
\affiliation{
National Key Laboratory of Parallel and Distributed Processing\\
Computer College, National University of Defense Technology
\country{China}
}
\affiliation{
Xiamen University
\country{China}
}
\email{sdiris@gmail.com}

\author{Xiangyu Ye}
\affiliation{
National Key Laboratory of Parallel and Distributed Processing\\
Computer College, National University of Defense Technology
\country{China}
}
\email{xyye@nudt.edu.cn}

\author{Yabo Duan}
\affiliation{
National Key Laboratory of Parallel and Distributed Processing\\
Computer College, National University of Defense Technology
\country{China}
}
\email{yaboduan@nudt.edu.cn}

%%
%% By default, the full list of authors will be used in the page
%% headers. Often, this list is too long, and will overlap
%% other information printed in the page headers. This command allows
%% the author to define a more concise list
%% of authors' names for this purpose.
% \renewcommand{\shortauthors}{Trovato and Tobin, et al.}

%%
%% The abstract is a short summary of the work to be presented in the
%% article.
\begin{abstract}

Distributed data-parallel training has been widely adopted for deep neural network (DNN) models. 
Although current deep learning (DL) frameworks scale well for \emph{dense} models like image classification models, we find that these DL frameworks have relatively low scalability for sparse models like natural language processing (NLP) models that have highly sparse embedding tables. Most existing works overlook the sparsity of model parameters thus suffering from significant but unnecessary communication overhead. In this paper, we propose \textit{EmbRace}, an efficient communication framework to accelerate communications of distributed training for sparse models. EmbRace introduces \textit{Sparsity-aware Hybrid Communication}, which integrates AlltoAll and model parallelism into data-parallel training, so as to reduce the communication overhead of highly sparse parameters. To effectively overlap sparse communication with both backward and forward computation, EmbRace further designs a \textit{2D Communication Scheduling} approach which optimizes the model computation procedure, relaxes the dependency of embeddings, and schedules the sparse communications of each embedding row with a priority queue. 
We have implemented a prototype of EmbRace based on PyTorch and Horovod, and conducted comprehensive evaluations with four representative NLP models. Experimental results show that EmbRace achieves up to 2.41$\times$ speedup compared to the state-of-the-art distributed training baselines.

\end{abstract}

%%
%% The code below is generated by the tool at http://dl.acm.org/ccs.cfm.
%% Please copy and paste the code instead of the example below.
%%


%%
%% Keywords. The author(s) should pick words that accurately describe
%% the work being presented. Separate the keywords with commas.
% \keywords{datasets, neural networks, gaze detection, text tagging}

%%
%% This command processes the author and affiliation and title
%% information and builds the first part of the formatted document.
\maketitle

\section{Introduction}
Recently, Deep neural networks (DNNs) have been extensively used in different domains like computer vision and natural language processing (NLP). 
\textit{Data parallelism} \cite{cite27} on distributed GPUs is the most widely used parallel strategy to reduce the overall training time. 
Unfortunately, with the fast increasing size of models and number of workers,
communication overhead becomes the main performance bottleneck of distributed training \cite{cite15}.

Communication acceleration has been extensively studied in the literature \cite{cite18,cite23,cite31}.
However, these approaches mainly focus on dense models like DNNs for computer vision. In sparse NLP models, embedding tables are widely-used for feature learning, holding a large portion of parameters. Only a subset of embedding is accessed and exchanged in one training step, bringing dramatic sparsity of embedding parameters in computing and communication, and making it a big challenge to scale the distributed training task efficiently. 

When treating all sparse parameters (tensors) as dense, it will involve unnecessary communication overhead and get poor scalability, especially on low-end GPUs. 
One line of work \cite{parallax,omni,hma,cite14} explores the sparsity-aware communication method, using an individual strategy to aggregate sparse gradients such as AllGather, Model Average and Parameter Server. 
However,
none of these works can reach high efficiency and scalability simultaneously, 
as the communication overhead of sparse aggregation is still non-negligible.

In addition to communication reduction,
recently many works focus on communication scheduling to overlap communication with computation \cite{p3,tic,pace,bytescheduler}. 
The key idea is to adjust the communication order of DNN layers to hide more gradient transmission by computation. 
However, existing scheduling approaches ignore the sparsity of models and the attributes of embedding tables, where model computation depends on the transmission of the entire embedding and the communication of embedding tables becomes the performance bottleneck.

To accelerate communication for distributed training of sparse models, 
in this paper we propose \textit{EmbRace}, a communication framework 
that integrates \textit{Sparsity-aware Hybrid Communication} with \textit{2D Communication Scheduling}.
Sparsity-aware hybrid Communication applies AlltoAll along with 
model parallelism
and combines AlltoAll and AllReduce together to race the gradients aggregation of sparse models. 
2D Communication Scheduling is introduced for effectively overlapping communication and computation by optimizing model computation procedure, employing fine-grained communication scheduling, and relaxing the computation dependency inside of embeddings.

We summarize our contributions as follows:

$\bullet$ We design and implement EmbRace, a distributed communication framework for sparse models atop PyTorch \cite{pytorch} and Horovod \cite{cite14}. We 
conduct AlltoAll along with column-wise partitioned model parallelism as our communication strategy for sparse embeddings, and analyze its efficiency. 
To the best of our knowledge, we are the first to bring model-parallel AlltoAll to data-parallel NLP training.

$\bullet$ We propose 2D Communication Scheduling which combines horizontal and vertical scheduling.
We dive into embedding matrices, calculate the minimum dependency of the next forward computation and arrange a communication schedule for each row of sparse gradients. This reduces the amount of communication that blocks the FP computation and enables forward computation to overlap more communication. 
As far as we know,
we are the first to realize communication scheduling inside sparse gradients.

$\bullet$ We comprehensively evaluate the performance of EmbRace using four representative NLP models on two kinds of GPU clusters with four popular distributed training methods. 
Experiments show that EmbRace accelerates the training process by up to 77.0\% on an RTX3090 GPU cluster and up to 141.5\% on an RTX2080 GPU cluster, compared to the fastest baseline.

\section{Background}

\subsection{Sparsity of NLP Models} \label{sec:modelsparsity}

A deep learning (DL) model is usually constructed by a sequence of layers and trained by repeating iterations. 
A typical training step involves three major parts: 1) \textit{forward pass} (\textit{FP}), 2) \textit{backward pass} (\textit{BP}) and 3) \textit{parameter update} with gradients from BP. In data parallelism, gradients are aggregated from all workers after BP to keep model synchronized, where network communication happens.

Most parameters of a DNN model are dense tensors represented by multi-dimensional arrays and stored contiguously in memory. But for sparse tensors full of zeros, only storing the non-zero elements is more efficient in memory footprint. In PyTorch, the coordinate (COO) format is the default sparse data storage format, which describes data as tuples of element indices and their corresponding values.

Embedding matrices are generally adopted in NLP models for feature learning, where each row maps a word to a vector. When training embeddings with batches, a small subset of vocabulary is used, and only the related rows in the embeddings would be updated. Therefore the gradients of embedding matrices could be represented in sparse format, and the communication of embedding gradients could also be sparse. 
As shown in Table \ref{tab:models}, popular NLP models have large portions of embedding parameters. These high proportions of embeddings hint that designing an individual communication scheme for embedding matrices in NLP models is necessary.

\begin{table}[tb]
    \caption{Model size and embedding size (MB) in popular NLP models. Ratio refers to the proportion of embedding parameters.}
  \label{tab:models}
  \centering
  \begin{tabular}{lccc}
    \hline

    Models &
    \begin{tabular}[c]{@{}c@{}}Model \\ Size \end{tabular}  & \begin{tabular}[c]{@{}c@{}}Embedding \\ Size \end{tabular} & 
    Ratio\\
    \hline
    LM \cite{lm} & 3186.5 & 3099.5	& 97.27\% \\
    GNMT-8 \cite{gnmt} &		739.1&	252.5&	34.16\% \\
    Transformer \cite{transformer} &	1067.5 &	263.4 &	24.67\% \\
    BERT-base \cite{bert} &		417.7 &	89.4 &	21.42\% \\
  \hline

\end{tabular}
\end{table}

\begin{figure}[t]
  \centering
  \subfigure[AllReduce]{
	\includegraphics[width=0.475\linewidth]{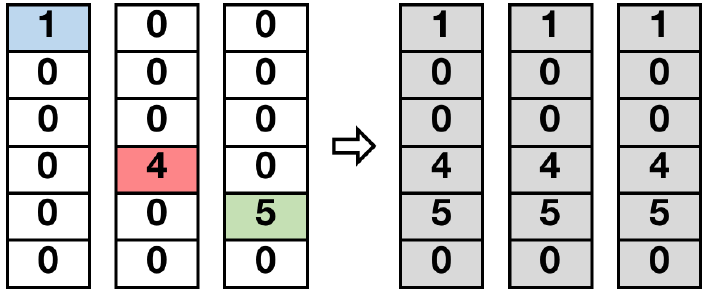}
  }
  \subfigure[AllGather]{
	\includegraphics[width=0.475\linewidth]{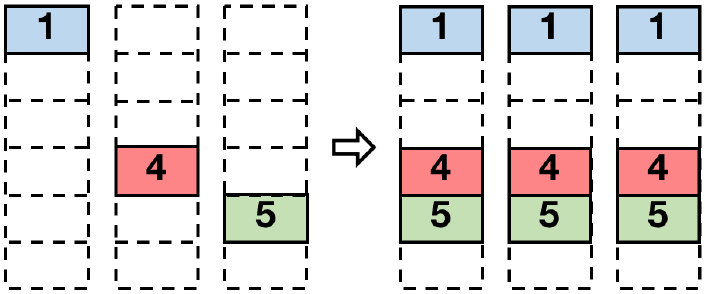}
  }
	\caption{Example sparse data transformation in AllReduce and AllGather with 3 processes.}
	\label{fig:sparsecomm}
\end{figure}

\subsection{Sparse Communication Architectures} \label{sec:commarchitects}

Due to the proven simplicity and high efficiency \cite{cite11,cite12}, recent DL frameworks employ collective routines as their communication methods and implement them on top of collective communication libraries such as MPI \cite{mpi}, NCCL \cite{nccl} and Gloo \cite{gloo}. 
In data-parallel training, all workers hold a model replica and synchronize gradients through specific collective primitives before each parameter update operation.
Two typical primitives are \textit{AllReduce} and \textit{AllGather}:

$\bullet$
AllReduce aggregates data from all processes, reduces the data with an operator such as sum, and distributes results back to each participant. In DNN training, AllReduce is 
widely used for aggregate gradients \cite{cite14} and 
implemented effectively \cite{cite1}.

$\bullet$
AllGather gathers the complete data from all workers and distributes the combined data to all workers. AllGather is practical when collecting non-associative tensors such as compressed gradients \cite{cite4, cite7}.

AllReduce is the most popular approach when researchers parallelize dense models such as image classification models \cite{cite18}. 
However, when distributing DL models which is sparse, AllReduce is no longer a suitable technique. How AllReduce and AllGather communicate sparse data on 3 processes are illustrated in Figure \ref{fig:sparsecomm}, AllReduce has to communicate and sum all data including zeros, while AllGather only sends the non-zero values.
Some training methods \cite{byteps} naively treat the sparse tensors in dense format to utilize the high-performance AllReduce implementation.
AllGather is naturally suitable to aggregate the sparse tensors thus popular distributed frameworks \cite{cite14,cite13} employ it for sparse communication.
We analyze this further in Section \ref{sec:spasrcomm}.

\subsection{Communication Scheduling}

During each BP procedure, the gradients generation depends on the computational graph of DNN. The gradients can start communication immediately after they are calculated rather than wait to complete the entire BP, which is suggested as wait-free backpropagation from Poseidon \cite{poseidon} and supported in most distributed DL frameworks \cite{cite13,cite14}.

In DL frameworks, computation and communication are typically carried out with a dependency directed acyclic graph (DAG).
In default DAG of DL frameworks, the order of gradients communication is based on a FIFO queue, and FP computations need to wait for the finish of all communications, making communication be overlapped only with the BP computation.
Thus comes the 
\textit{communication scheduling} \cite{p3,bytescheduler}, 
which switches the FIFO queue into a priority queue to adjust communication orders. 
The gradient communications that are blocking the beginning of the next FP would be prioritized.
So that FP could start earlier and both BP and FP could overlap communication operations.

\section{Challenges and Motivations} \label{sec:motivation}
The sparse and massive embedding tables in NLP models bring serious challenges to efficient communication and scalable distributed training. Previous works pay little attention to this problem, or could not achieve satisfying results. 
Deeply theoretical and practical analysis of embedding tables motivates us to work on a scalable communication framework for distributed NLP model training.     

\textbf{Efficient communication for embedding tables.}
Embedding tables occupy an important role in NLP models along with massive parameters, which introduce large communication overhead and seriously restrict the training scalability.
Some approaches use hybrid communication architectures combining PS \cite{parallax} or model average \cite{hma} with AllReduce. OmniReduce \cite{omni} optimizes AllReduce algorithm to suit the sparse situation. 
However, based on our analysis in Section \ref{sec:spasrcomm} and experience, these solutions are still sub-optimal.
Hence, communicating embedding in NLP models with both high efficiency and good scalability, becomes a challenge for us.

\begin{figure}[tb]
    
    \subfigure[Scheduling all communication in dense format.]{\includegraphics[width=0.85\linewidth]{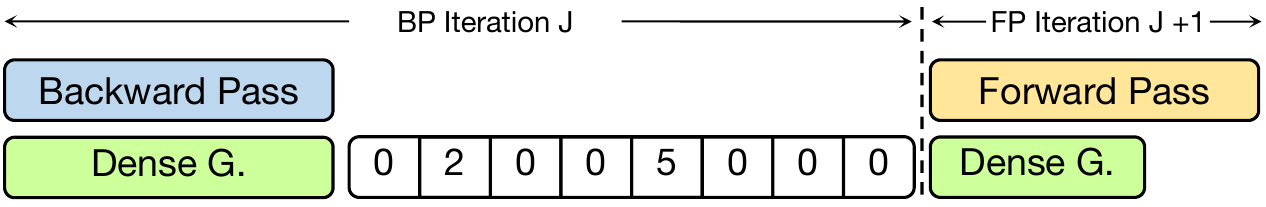}}
    
    \subfigure[2D Communication Scheduling, our fine-grained communication scheduling approach for each token of embedding tables.]{\includegraphics[width=0.85\linewidth]{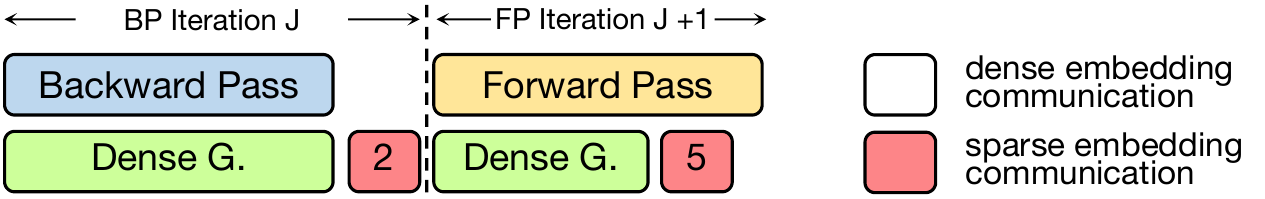}}

	\caption{Execution timeline comparisons between communication scheduling approaches, Dense G. refers to the communication operation of dense gradients.}
	\label{fig:schedulecompare}
\end{figure}

\textbf{Scheduling for sparse communication.}
Overlapping communication with computation could reduce the communication overhead and accelerate the distributed training. Researchers use communication scheduling \cite{p3,tic,bytescheduler,pace} to achieve this purpose.
However, to the best of our knowledge, existing scheduling approaches are only designed for dense models, treating embedding tables as an entire scheduling unit. Embedding tables in dense format will involve unnecessary communication overhead and prevent the starting of forward computation. As shown in Figure \ref{fig:schedulecompare}(a), the communication of zero elements in Embedding would slow the execution of FP.
Moreover, embeddings have a unique attribute: parameter updates of embedding tables are element-wise, relating to the tokens which are appeared in a training batch. 
Based on this property, as illustrated in Figure \ref{fig:schedulecompare}(b), developing an exhaustive scheduling method for NLP tasks to overlap more communication with computation, becomes another challenge for us.

\begin{figure}[tb]
  \centering
    \includegraphics[width=1\linewidth]{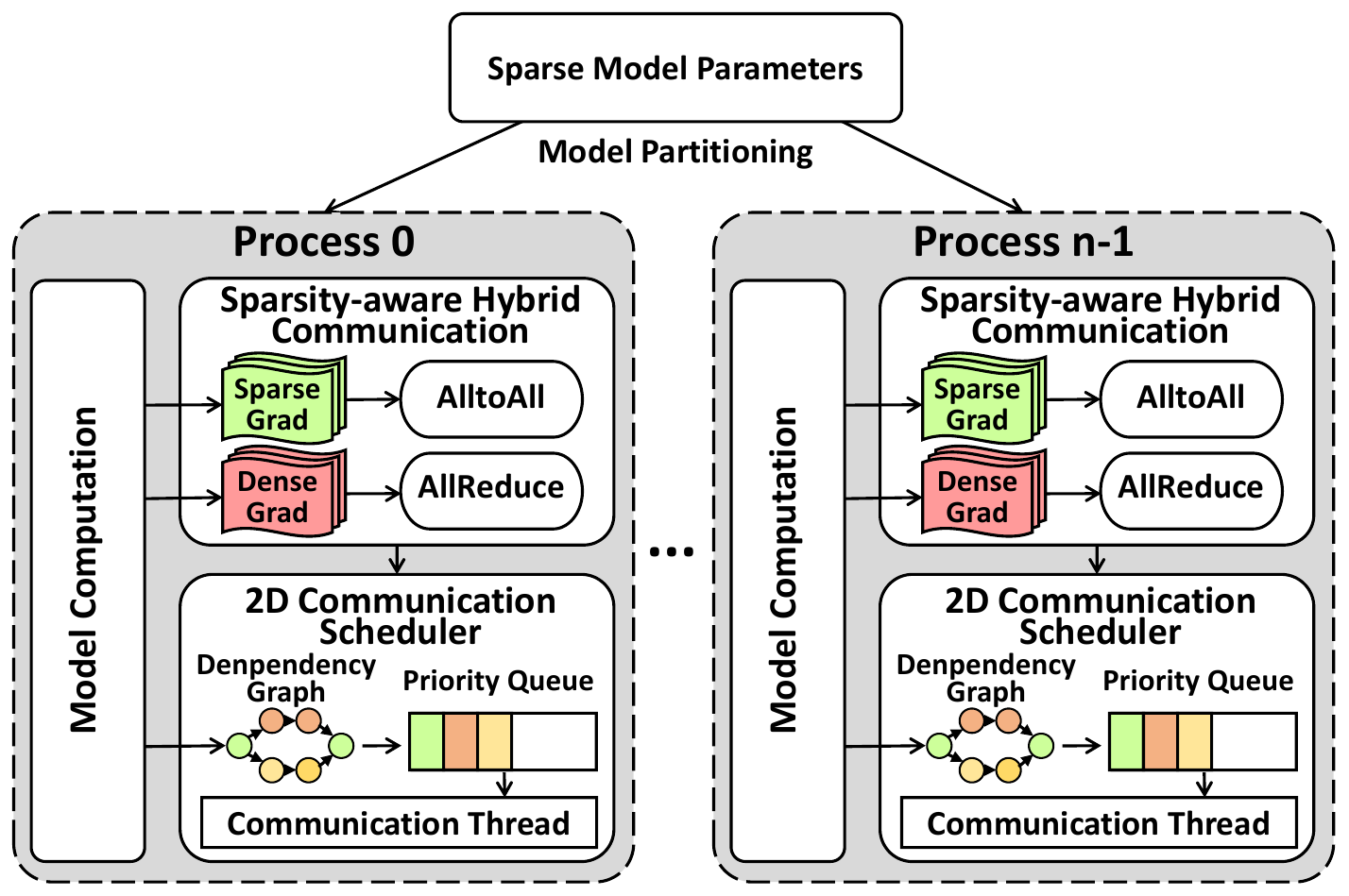}
    \caption{Overview of EmbRace, an efficient communication framework with Sparsity-aware Hybrid Communication and 2D Communication Scheduling.}
    \label{fig:overview}
\end{figure}

\section{EmbRace Design}
Motivated by Section \ref{sec:motivation}, we design EmbRace to address the challenges. Figure \ref{fig:overview} shows an overview of EmbRace. 
To achieve efficient communication for both sparse and dense data in NLP models, we design Sparsity-aware Hybrid Communication, applying hybrid communication strategies. For sparse tensors, we apply model parallel and AlltoAll to specifically speed up their communications. For dense tensors, AllReduce will take charge as usual.
To efficiently schedule the communication of sparse NLP models, we introduce 2D Communication Scheduling. 
With the help of dependency graph calculation, we could assign priorities to all dense and each block of sparse gradients in model and hold a priority queue to perform all communications operations.

\subsection{Sparsity-aware Hybrid Communication}
\subsubsection{AlltoAll with Model Parallelism}
In various NLP tasks, sparse tensors come with embedding tables. 
As suggested in Section \ref{sec:commarchitects}, there are potential benefits from applying an individual communication scheme for sparse tensors.

Collective operation AlltoAll is useful in recommender system training \cite{cite5}, where processes transmit and receive data slices from every other process to redistribute the data. We bring it with model parallelism to data-parallel NLP training for pursuing the efficiency of sparse communication.

We partition embedding tables for the AlltoAll primitive at first.
There are two typical partition solutions, row-wise partitioning and column-wise partitioning.
Since each row represents a word in embedding tables, the row-wise partitioning will split different words along with their complete vectors. But the word frequencies are distinct in most datasets, some partitions will be accessed much more frequently, leading to an unbalancing communication cost. 
In contrast, the column-wise partitioning method will scatter the embedding vectors, keeping a whole corpus vocabulary in each part. The uniform partitions will get the same amount of requests and own balance loads naturally.

The full embedding table is column-wise partitioned into processes before the training start. In each iteration, embedding in each process firstly looks up all training data of this step and produces a different embedding result. Then AlltoAll is called for redistributing the embedding results so that each process gets one embedding result batch and injects it into dense modules. Next, each process computes sparse gradients of embedding tables. Finally, AlltoAll is called again to exchange sparse gradients between processes and each process could update embedding parameters with new gradients.

\subsubsection{Efficiency Analyses} \label{sec:spasrcomm}

Firstly we prove the efficiency of our implement theoretically.
Suppose we train an NLP model in a cluster with $n$ nodes, and each node is equipped with $w$ GPU workers. $N$ denotes the total GPU number and $N=w\times n $. There is one embedding table with size $M$ in the model. In each training step, the average density of embedding gradients is $\alpha$ and its sparsity will be $1-\alpha$, which means the gradient size could be approximated to $M$ when stored as a dense tensor and $\alpha M$ as a sparse tensor. To simplify, we assume that bandwidths $B$ and communication start latency $\beta$ between any two workers are uniform.

With \textbf{AlltoAll}, embedding tables are partitioned among $N$ GPUs. In each training step, AlltoAll is performed twice, one for redistributing embedding lookup results and another for aggregating embedding gradients.
A single AlltoAll operation will involve data exchanges with other $ N-1 $ training processes, the communication amount of each exchange will be $ \frac{\alpha M}{N}$ and the time cost will be $ \frac{\alpha M}{NB} + \beta $.
Hence the overall communication overhead will be $2 (N-1)(\frac{\alpha M}{NB} + \beta)$.

\textbf{AllReduce}, which does not support sparse tensor, limits the gradient format to dense. In the most widely used ring-based AllReduce, each GPU divides the gradient into $N$ parts and the communication unit size is $\frac{M}{N}$. Then workers use a ring topology to get the summation of parts by sending and receiving a gradient unit from neighbors for $N-1$ times. Then, the summed results are gathered through the same ring for another $N-1$ times. Therefore the overall communication cost is $2(N-1)(\frac{M}{NB}+\beta)$.

With \textbf{Parameter Server (PS)} \cite{ps} architecture, we represent the number of servers by  $S$, which holds $ S \leq n$. Since we partition the embedding parameters equally across servers, the message size will be $ \frac{\alpha M}{S}$ each. During each iteration, servers send parameters to GPUs and receive relative gradients. Both operations transmit message size of $N \times\frac{ \alpha  M}{S}$, so that the total overhead will be  $ 2N ( \frac{\alpha M}{SB}+\beta) $, whose lower bound will be $  2N ( \frac{\alpha M}{nB}+\beta)$.

In \textbf{AllGather}, GPU workers send and receive the gradient in sparse format to each other simultaneously. Since the number of workers is $N$, the data communication will repeat for $(N-1)$ times with the gradient size of $\alpha M$. After summing up the transmitted data size to $(N-1)\alpha M$,  the time will be $(N-1)( \frac{\alpha M}{B} + \beta)$.

Table \ref{tab:commcost} summaries the communication costs. 
Comparing the estimated time results, the AlltoAll, AllReduce, and PS maintain good scalability, where communication will not be strongly impacted by GPU number $N$.
In distributed training environments with $N >1 $ and $N \ge n$, for sparse tensors which hold $\alpha \leq 1 $, the AlltoAll method would be faster than AllReduce and PS theoretically. 
Although in some rare situations with a small number of $N$ and long starting latency $\beta$, AlltoAll is slower than AllGather, the transmissions time of AllGather is approximately linear to the GPUs number $N$ with poor scalability.

\begin{figure}[tb]
    \centering
    \subfigure[2 nodes with 4 RTX3090 GPU each]{
        \includegraphics[width=0.475\linewidth]{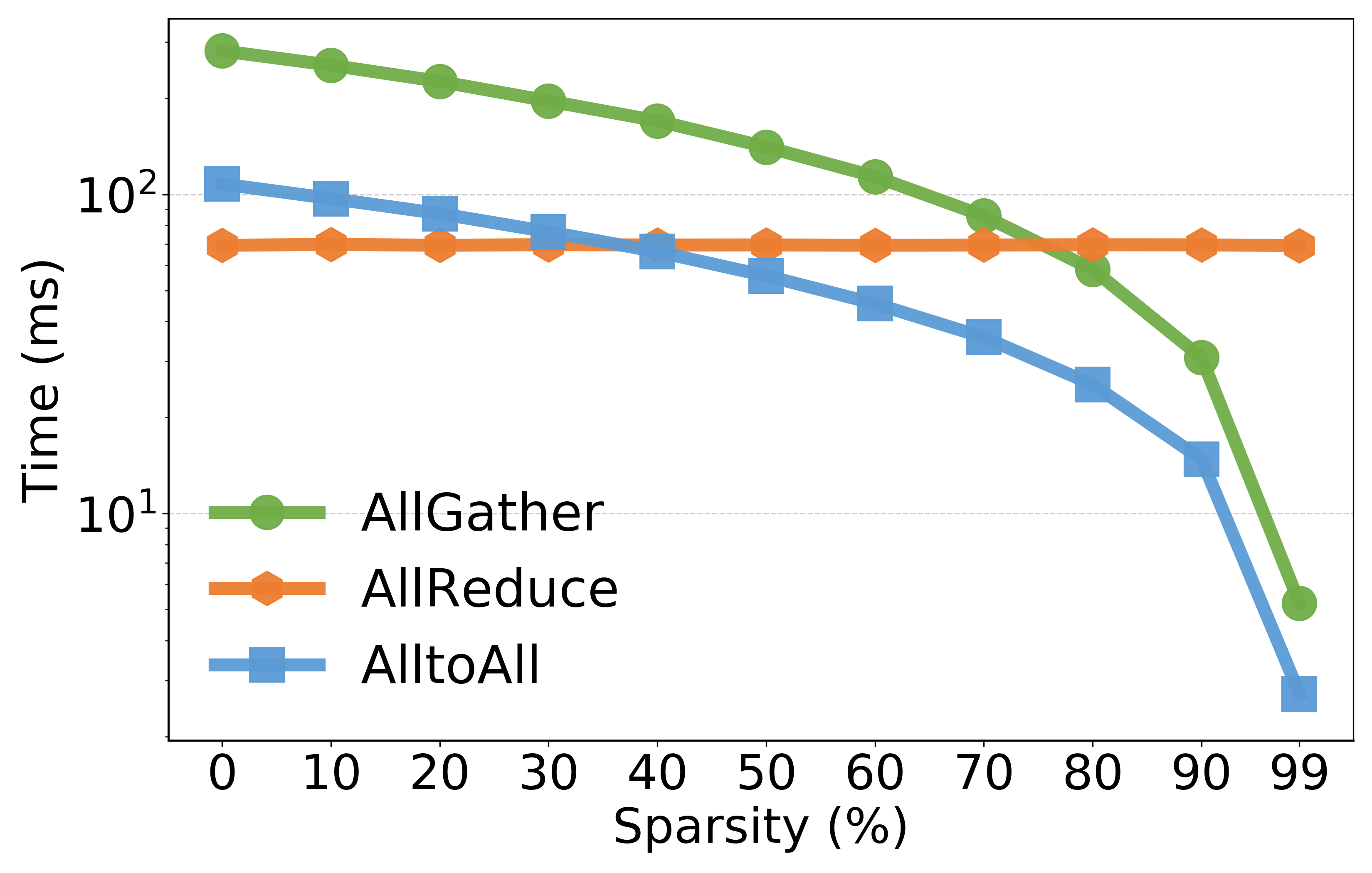}
    }  
    \subfigure[4 nodes with 1 RTX3090 GPU each]{
        \includegraphics[width=0.475\linewidth]{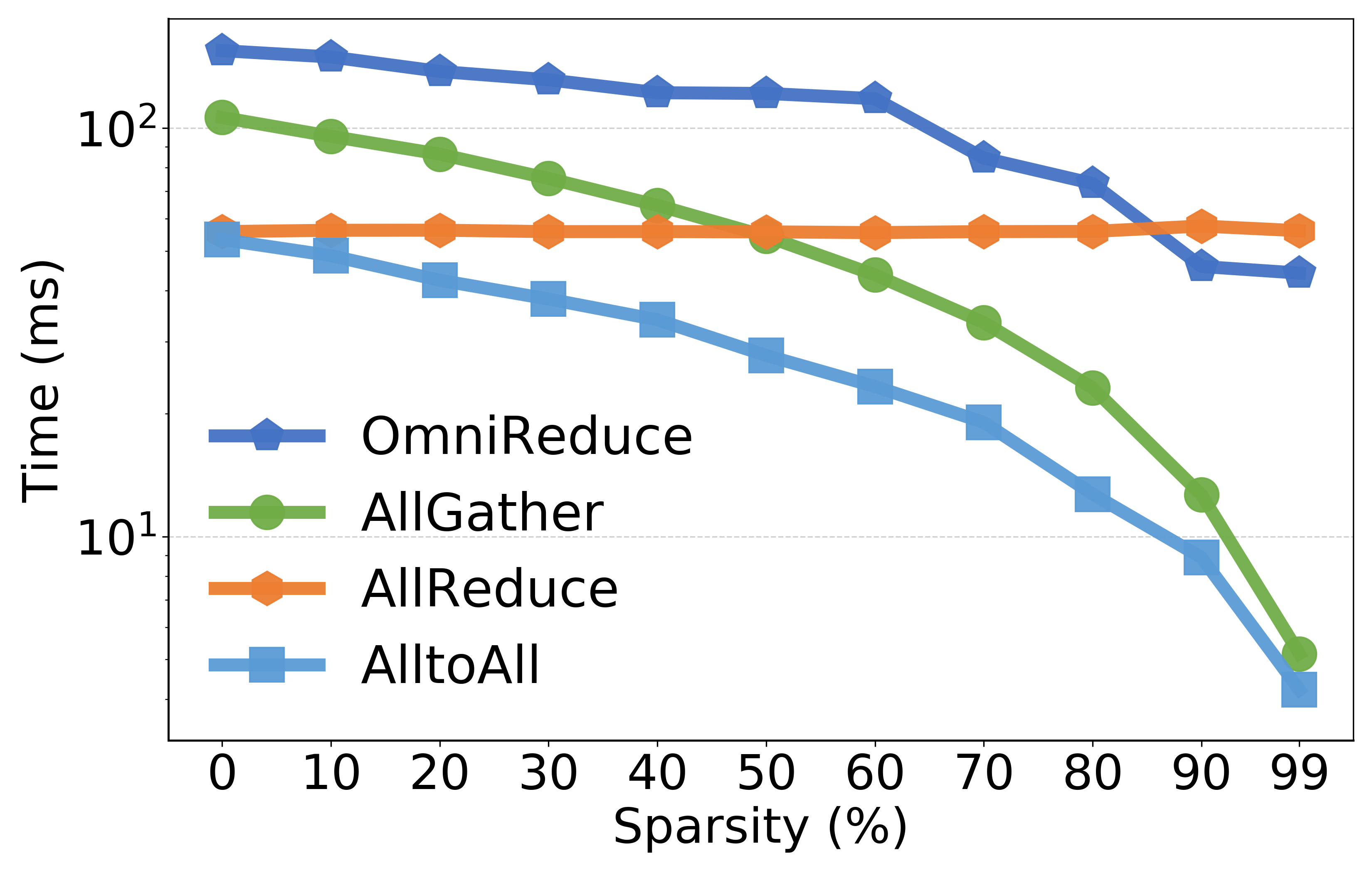}
    }    
    \caption{Example communication overheads of embedding gradient with different sparsity and communication schemes. Embedding comes from GNMT-8 model and its size is 252.5 MB. OmniReduce \cite{omni} implements a sparsity-aware AllReduce algorithm but only supports each node uses 1 GPU.}
\label{fig:sparsity}
\end{figure}

However, in the practical training scenario, different communication algorithms, network topologies and message sizes would influence the bandwidth utilization greatly. Figure \ref{fig:sparsity}a shows an example of communication overheads on 8 RTX3090 GPUs, AlltoAll outperforms other methods when the sparsity is greater than 40\%. Take models of Table \ref{tab:models} for example, when batch size per worker of LM, GNMT-8,
Transformer and BERT-base are 128, 128, 5120 and 32, their corresponding average sparsity are 99.7\%, 89.7\%, 86.6\% and 59.7\%.
Figure \ref{fig:sparsity}b shows another case on 4 RTX3090 GPUs across 4 server nodes, AlltoAll is the best method in all sparsity. OmniReduce could reduce the communication overheads along with the increase of sparsity, but they suffer from insufficient bandwidth usage with excessive divided messages.
Therefore, after considering both theoretical and practical results, AlltoAll becomes our sparse communication solution.

\begin{table}[tb]
\renewcommand\arraystretch{1.2}
\renewcommand\tabcolsep{12.0pt}
    \caption{Communication overhead of a sparse tensor according to the communication approaches.}
  \label{tab:commcost}
  \centering
  \begin{tabular}{lr}
    \hline
    Approaches & Communication Overhead \\
    \hline
    AlltoAll&  $2 (N-1)(\frac{\alpha M}{NB} + \beta)$ \\
    AllReduce & $2(N-1)(\frac{M}{NB}+\beta)$ \\
    PS &  $  2N ( \frac{\alpha M}{nB}+\beta)$\\
    AllGather & $(N-1)( \frac{\alpha M}{B} + \beta)$\\
  \hline
\end{tabular}
\end{table}

\subsubsection{Hybrid Communication Architecture}

Apart from sparse data, we employ AllReduce architecture to aggregate dense gradients where $\alpha = 1$. The dense parts are treated as an individual dense model so that we could utilize the popular distributed DL framework Horovod.

In summary, we implement a hybrid communication architecture, 
which combines two collective primitives AllReduce and AlltoAll, and associates data parallelism with model parallelism.
We use AlltoAll for sparse communications of embedding tables with column-wise partitioning and prove its efficiency.

\subsection{2D Communication Scheduling}
To further accelerate distributed training among GPUs, we focus on two main points: reducing the computation stall by overlapping communication with computation; and reducing the communication amount by utilizing the GPUs' idle time.
To deeply overlapping, 
we firstly schedule dense blocks and FP of embedding layers with a priority queue according to the dependency graph. 
These make our Block-level Horizontal Scheduling.
Furthermore, 
we step into the embedding matrix to reduce the communication quantity. We calculate the minimal gradients dependency, select the necessary gradient rows for the next iteration, prioritize their communication, and delay the remaining communication. These make our Vertical Sparse Scheduling.
Combing the horizontal and vertical scheduling, we propose our 2D Communication Scheduling for NLP tasks.

\subsubsection{Block-level Horizontal Scheduling}

\begin{figure}[tb]
  \centering
    \includegraphics[width=1\linewidth]{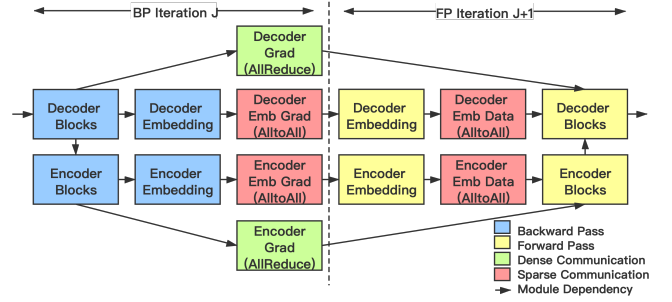}
    \caption{Module dependency graph with Sparsity-aware Hybrid Communication in translation tasks starting from the backward pass. The \textit{Emb Grad} and \textit{Emb Data} refer to the sparse gradients and sparse lookup results of the embedding table.}
    \label{fig:dependency}
\end{figure}

\begin{figure*}[tb]
  \centering
    \subfigure[Default Scheduling. Communication operations follow a FIFO queue, and communications are only overlapped with BP computation. It's the default method in popular DL frameworks like PyTorch and TensorFlow.] {
    \includegraphics[width=0.98\linewidth]{figures/timeline_a.pdf}
    }
    
    \subfigure[Block-level Horizontal Scheduling. Communications follow a priority queue, so that dense gradient transmissions, i.e. the green Decoder Blocks and Encoder Blocks, are split into parts, and some parts are overlapped with FP computation.]{
    \includegraphics[width=0.98\linewidth]{figures/timeline_b.pdf}
    }
    
    \subfigure[2D Scheduling. With Vertical Sparse Scheduling computation, the embedding gradients size required by FP is reduced, hence the sparse communication overhead before FP is reduced. The effects of Vertical Sparse Scheduling (Algorithm \ref{alg:schedule}) are shown in the right dotted grid.]{
    \includegraphics[width=0.98\linewidth]{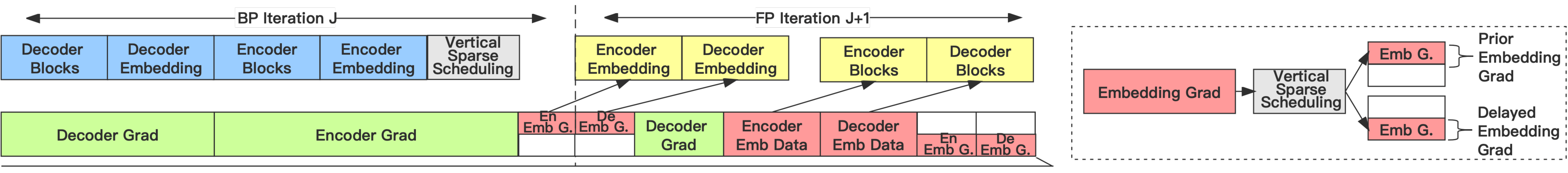}
    }
    \caption{Example execution timelines starting from the backward pass of EmbRace with different communication scheduling schemes. The \textit{Emb Grad (G.)} and \textit{Emb Data} refer to the corresponding sparse gradients and sparse lookup results of \textit{Encoder (En)} or \textit{Decoder (De) Embedding}.}
    \label{fig:timeline}
\end{figure*}

In order to schedule NLP models properly, we explore the computation graph dependency at first.
Taking translation models (e.g., Transformer, GNMT-8) as examples, they can be broken down into the following major modules:
1) \textit{Encoder Embedding}; 2) \textit{Encoder Blocks}; 3) \textit{Decoder Embedding}; 4) \textit{Decoder Blocks}.
With our Sparsity-aware Hybrid Communication scheme, the dependencies between FP, BP and communication are shown in Figure \ref{fig:dependency}.  

In general, the orders of FP and BP are inverse, and the communication would follow a FIFO queue by default.
The finish of all gradient communications would prevent starting the FP of the next iteration, and the FP of Encoder and Decoder Blocks have to wait for their related communicated embedding lookup results. Figure \ref{fig:timeline}(a) depicts the execution timeline of computation and communication with FIFO situation, which is adopted by most distributed frameworks.

To reduce the waiting time, we adopt a priority queue.
Unlike popular scheduling methods which use tensor partitioning and involve two inefficiency: extra communication starting overhead due to the increasing number of communication operations; inadequate bandwidth utilization due to the small partitioned message size. 
We assign the priority according to the following rules.
For dense modules of NLP tasks, we observe that the computation loads are more even than image classification models. For example, there are 12 self-attention blocks in Bert-base encoder, each holds a similar number of parameters and takes a comparable calculation time. This similarity hints at applying scheduling with these blocks, where parameters in the same block got the same priority and transmit their gradients together. 
The dense blocks get priority according to the FP dependency order so that their FP could start as soon as communications finish.

Another characteristic of NLP tasks is that the FP of embedding layers neither depends on each other nor other forward passes. 
We could perform embedding FP in advance and delay the FP of Encoder Blocks. This running sequence could make room for the AlltoAll communication of embedding data and rescheduled AllReduce communication of dense gradients.

Figure \ref{fig:timeline}(b) illustrates an example execution timeline of Block-level Horizontal Scheduling.
Compared to the timeline with default scheduling in Figure \ref{fig:timeline}(a), the communication quantity is the same, but a large amount of communication is overlapped with FP computation. 
However, limited by bandwidth and module dependency, there is some stalling after BP and some hang between FP.

\subsubsection{Vertical Sparse Scheduling}
Besides the order adjustment, there are potential reductions in embedding sparse gradients size from the following observations.

\textbf{Coalescing Gradients.}
The data batches in NLP are often generated by sentences.
On the one hand, there are multiple duplicate words in a single sentence naturally. On the other hand, when a tokenizer \cite{cite19} deals with sentences into uniformly shaped batches, the same value will be padded. 
With padding and duplicate words, the sparse embedding gradients would have repeated coordinates in the indices. These multi-valued elements could be coalesced into a single value using summation and hence reduce the gradient size. The reduction effect among models is shown in the coalesced size column of Table \ref{tab:gradsize}. In LM, GNMT-8, Transformer and BERT-base, the average gradient size is reduced by 20.4\%, 53.1\%, 52.9\% and 84.7\%, respectively.

\begin{table}[b]
    \caption{Average sparse embedding gradient size (MB) in Vertical Sparse Scheduling. The batch size per worker of LM, GNMT-8, Transformer and BERT-base are 128, 128, 5120 and 32.}
  \label{tab:gradsize}
  \centering
  \begin{tabular}{lccc}
    \hline
    Models & 
    \begin{tabular}[r]{@{}r@{}}Original \\ Grad Size \end{tabular}  & \begin{tabular}[r]{@{}r@{}}Coalesced \\ Grad Size \end{tabular} & \begin{tabular}[r]{@{}r@{}}Prioritized \\ Grad Size \end{tabular} \\
    \hline
    LM  & 8.7 & 6.9 & 2.6 \\
    GNMT-8 & 26.0 & 12.2 & 5.8 \\
    Transformer& 35.2 & 16.6 & 8.9 \\
    BERT-base& 36.0 & 5.5 & 3.2 \\
  \hline
\end{tabular}
\end{table}

\textbf{Minimum Dependency of Embedding.}
Vocabulary is usually much larger than each batch, leading to that only a small subset of embedding is used in each iteration. With the data changing among training steps, different corresponding embedding rows would be updated. This shifting prompts us that the minimum dependency of the subsequent embedding FP is the up-to-date embedding of the following batch data rather than a fully updated embedding.
Hence we adopt the data prefetch technology, which always keeps the data of the next iteration in memory.
Thanks to the prefetch, we are aware of the data used in the next iteration. 
When we consider communicating sparse gradients in this iteration, we divide the gradients into two parts, a necessary part and an unhurried part. The necessary part contains the gradients related to the intersection of data between the current and next iteration. In contrast, remaining gradients compose the unhurried part, which could be delayed.
Moreover, we assign the highest communication priority to the necessary part and the latest to the unhurried part, making them \textit{prior gradients} and \textit{delayed gradients}. The prioritized gradient size column of Table \ref{tab:gradsize} shows the average size of prior gradients. Compared with the coalesced gradient size, the gradient size is further dropped by 61.8\% in LM, 52.5\% in GNMT-8, 46.3\% in Transformer and 41.9\% in BERT-base.

\begin{algorithm}[tb]
\small
\caption{Vertical Sparse Scheduling}
\label{alg:schedule}
\DontPrintSemicolon
\SetAlgoNoLine 
  \KwIn{sparse gradient $G$, gathered training data for current iteration $D_{cur}$, gathered training data for next iteration $D_{next}$, process rank $n$}
  \KwOut{prior sparse gradient $G_p$, delayed sparse gradient $G_d$} 
  \SetKwProg{Init}{After BP: }{}{}
  \Init{}{
      \tcc{Coalesce the duplicate rows}
      $G_{coalesced} \gets COALESCE(G)$ \\
      \tcc{Get the unique training data of process rank $n$}
      $D_{u} \gets UNIQUE(D_{cur}[n])$ \\
      $i_{prior} \gets D_{u} \cap  D_{next}$ \\
      $i_{delayed} \gets D_{u} \setminus i_{prior}$ \\
      \tcc{Select the prior and delayed gradients from coalesced gradients}
      $G_p \gets INDEX\_SELECT(G_{coalesced}, i_{prior}) $ \\
      $G_d \gets INDEX\_SELECT(G_{coalesced}, i_{delayed}) $ \\
  }
  return \{$G_p,\ G_d$\} 
\end{algorithm}

\textbf{Vertical Scheduling.}
The cutting down and scheduling will modify the number of rows in sparse embedding gradients. The deciding algorithm is shown in Algorithm \ref{alg:schedule}, line 2-3 represent the computation of coalescing gradients, the line 4-7 refers to a sequence of set operations to distinguish prior gradients. The calculations require a considerable computing resource, and the GPU idle time after BP is a good occasion. The timeline after 2D Communication Scheduling is illustrated in Figure \ref{fig:timeline}(c): the grey box represents the Vertical Sparse Scheduling, whose impact is elucidated at the left dotted grid.
The embedding gradient sizes are reduced, and the sparse communication is done in two parts. The communications of prior gradients must be finished before embedding FP and the communications of delayed gradients could be performed later. The embedding FP starts immediately after communicating the slight prior gradients, resulting in better overlapped and efficient data-parallel training.

\begin{figure*}[tb]
\centering
\subfigure[LM, RTX3090, 1.18$\times$-1.77$\times$]{
\includegraphics[width=0.235\linewidth]{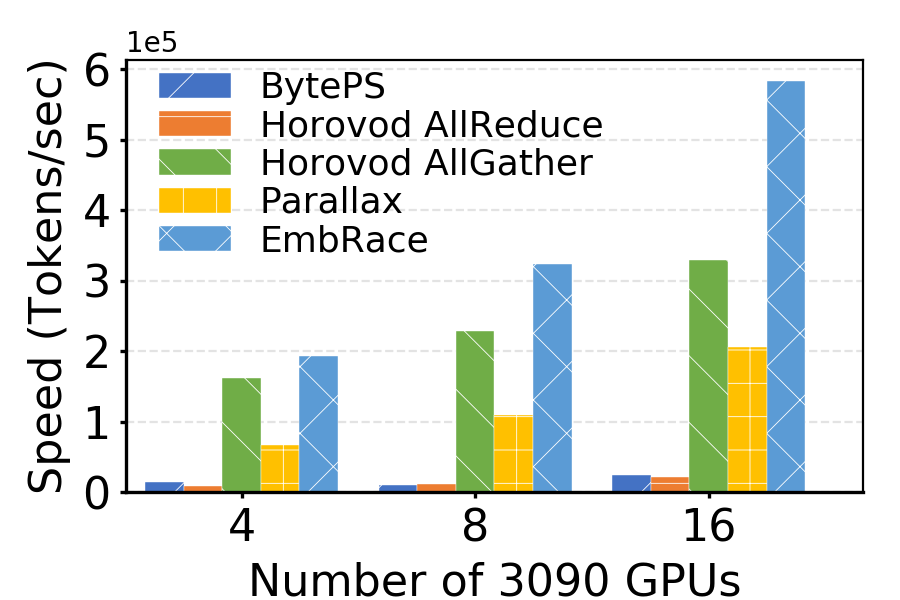}
}
\subfigure[GNMT-8, RTX3090, 1.10$\times$-1.27$\times$]{
\includegraphics[width=0.235\linewidth]{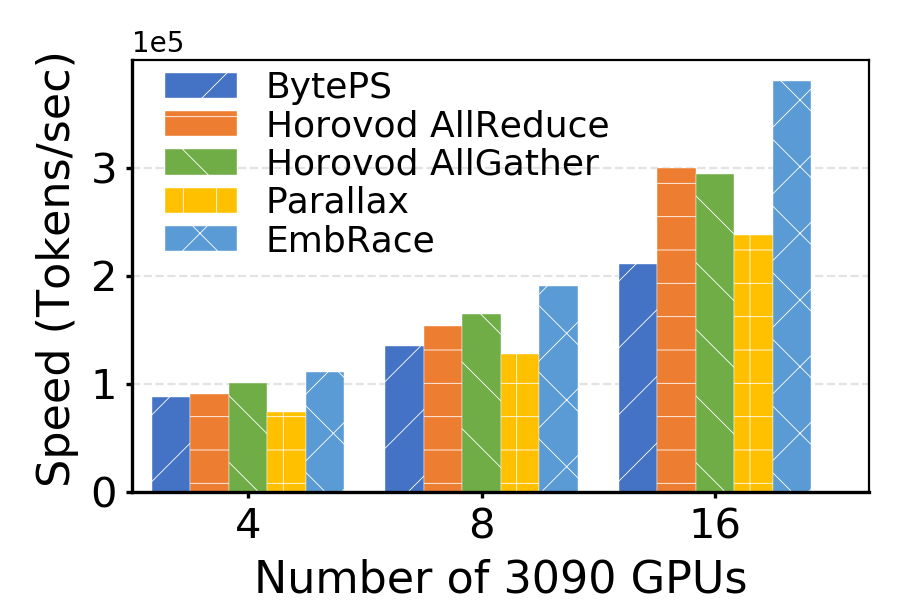}
}
\subfigure[Transformer, RTX3090, 1.12$\times$-1.18$\times$]{
\includegraphics[width=0.235\linewidth]{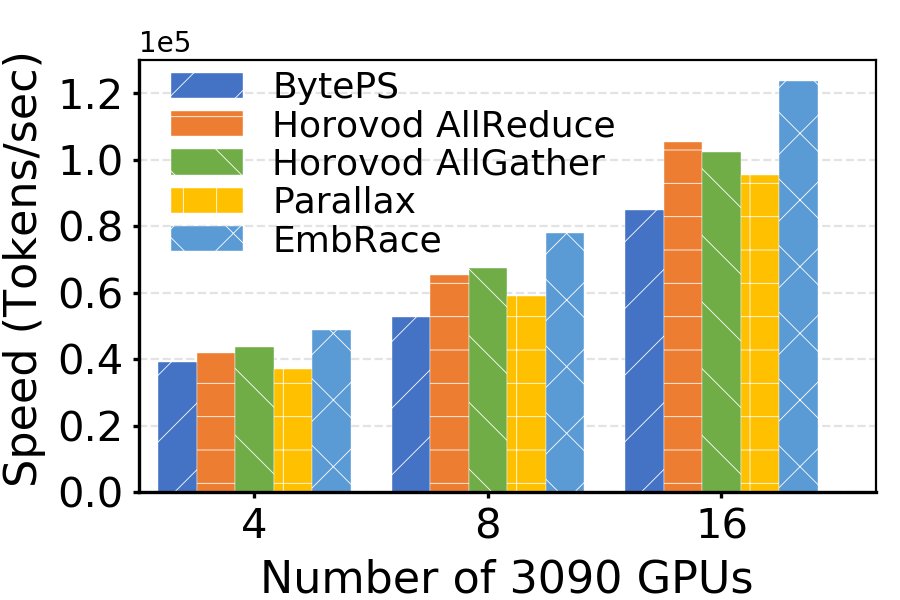}
}
\subfigure[BERT-base, RTX3090, 1.02$\times$-1.06$\times$]{
\includegraphics[width=0.235\linewidth]{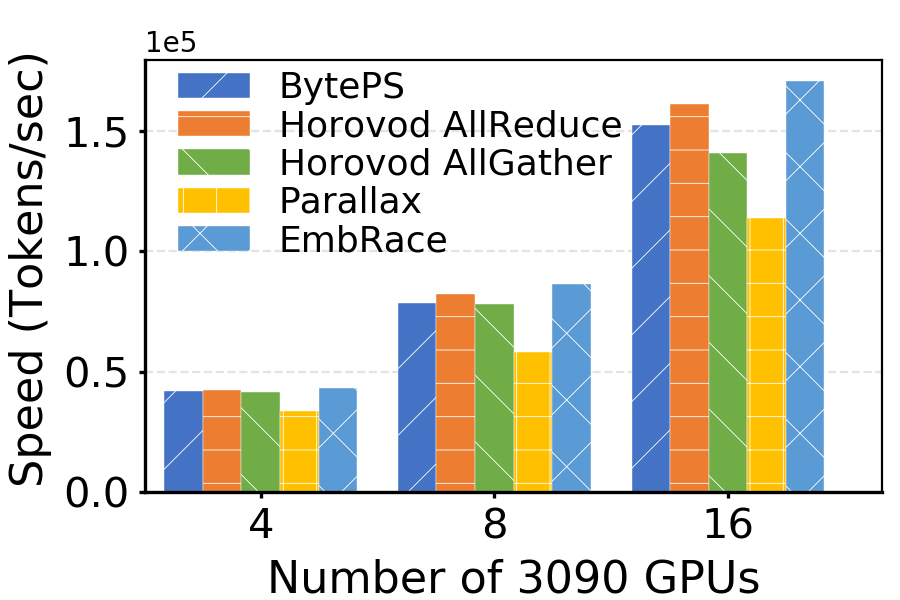}
}

\subfigure[LM, RTX2080, 1.99$\times$-2.41$\times$]{
\includegraphics[width=0.235\linewidth]{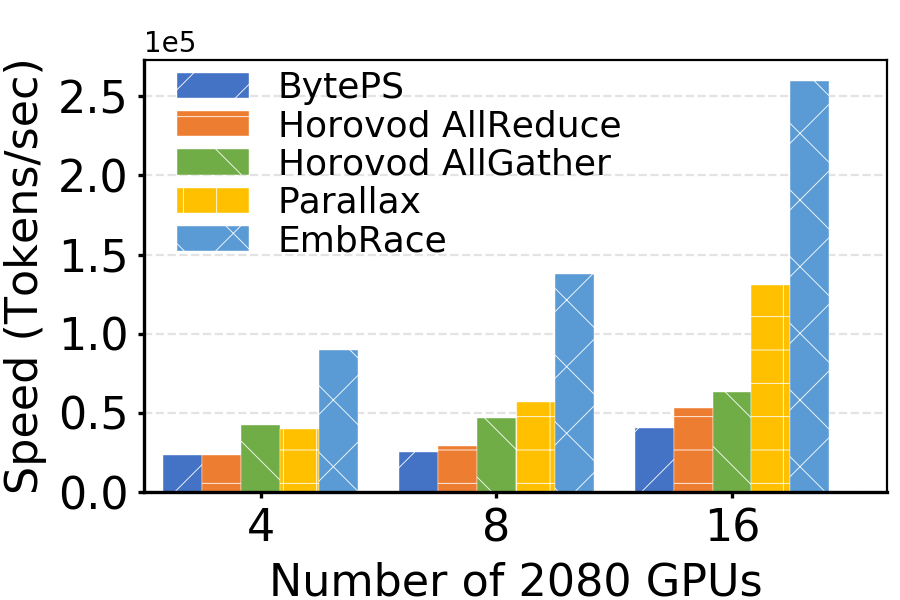}
}
\subfigure[GNMT-8, RTX2080, 1.09$\times$-1.30$\times$]{
\includegraphics[width=0.235\linewidth]{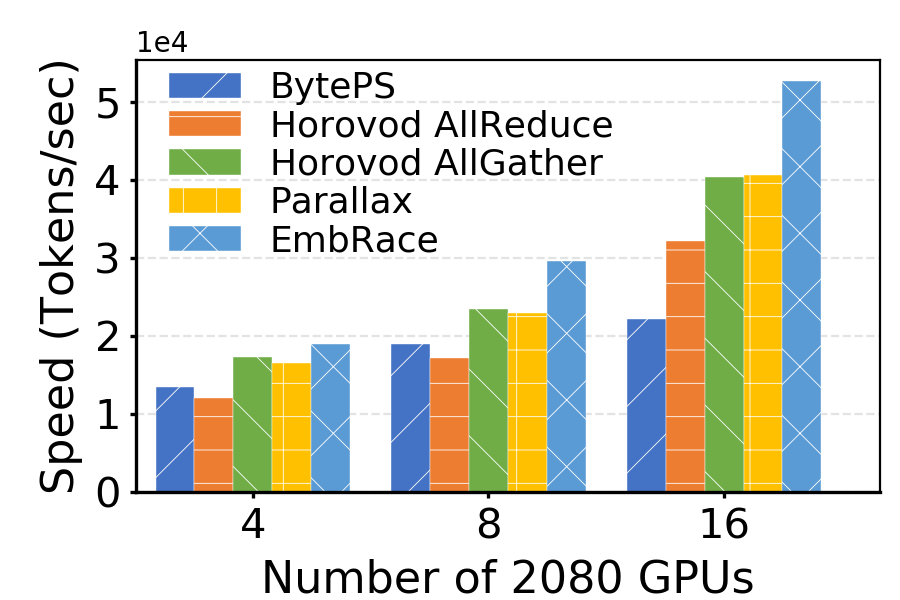}
}
\subfigure[Transformer, RTX2080, 1.11$\times$-1.28$\times$]{
\includegraphics[width=0.235\linewidth]{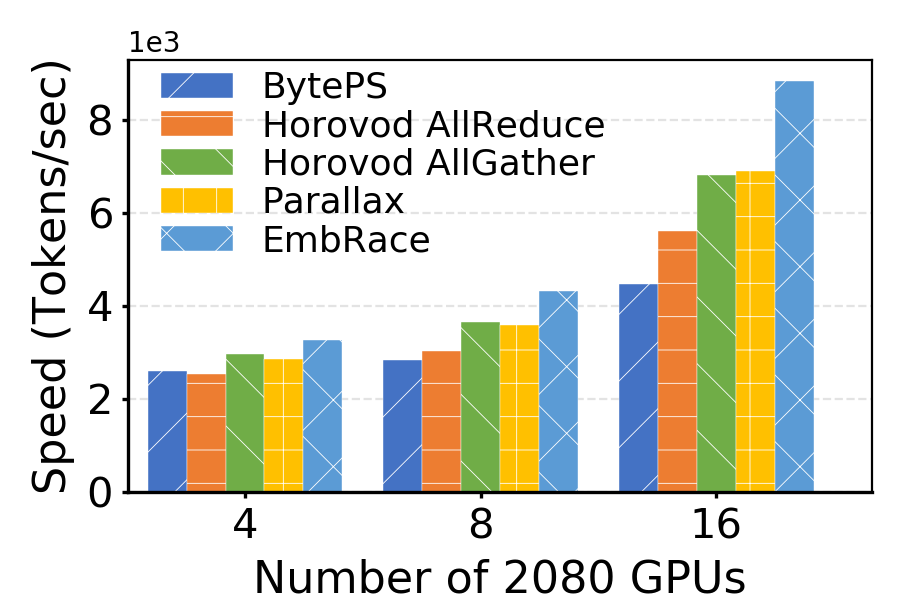}
}
\subfigure[BERT-base, RTX2080, 1.10$\times$-1.40$\times$]{
\includegraphics[width=0.235\linewidth]{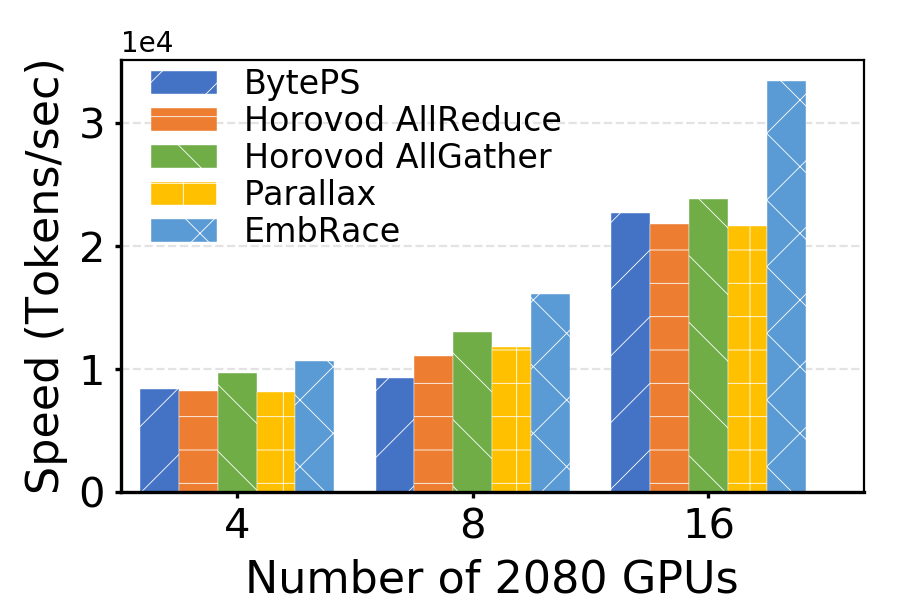}
}
\caption{The training performance of NLP models using different numbers of GPUs. The numbers are speedups of EmbRace over the best baseline.}
\label{fig:performance}
\end{figure*}

\section{Evaluations} \label{sec:eval}
\subsection{Implementation}
We implement EmbRace on top of DL framework PyTorch 1.8 along with distributed training framework Horovod 0.21.3.
EmbRace adopts the open-source communication library NCCL 2.7.8 as the collective routines executor.

EmbRace is integrated with Horovod for usage convenience but takes control of the communication operations. Apart from the same additions of Horovod, EmbRace only requires a few extra lines of code changes in PyTorch to prefetch data and reschedule the FP order. 
To carry out the Sparsity-aware Hybrid Communication, we first partition the embeddings according to the number of training processes. Then, we replace the AllReduce operations of word embedding gradients with AlltoAll, and we add another AlltoAll to collect embedding FP outputs by registering hooks.
As for 2D Communication Scheduling, we hold a priority queue and a communication thread. Communications are performed in the communication thread according to the priority queue. For the Block-level Horizontal Scheduling, before training starts, we assign a priority to each dense block according to the dependency graph and register a hook on each BP of dense blocks. When this hook is fired, the corresponding dense communication operations along with their priorities are dumped into our priority queue. For the Vertical Sparse Scheduling computation, we register another hook on the last BP to finish the calculation, assign corresponding priorities to computation results and add their communications into the priority queue as well.

\subsection{Experiments Setup}

\subsubsection{Hardware Configurations} 
Two 16-GPU clusters are used in our experimental evaluation. 
One type of server is equipped with four NVIDIA GeForce RTX3090 GPUs with 24GB GPU memory and six 16G DDR4 RAMs;
another has four NVIDIA GeForce RTX2080 GPUs with 8GB GPU memory and three 32G DDR4 RAMs.
All servers have two Intel Xeon 4214R @ 2.40GHz CPUs, run 64-bit Ubuntu 18.04, CUDA 11.1, cuDNN 8.0.5 and are connected by 100 Gbps InfiniBand.

\subsubsection{Models and Datasets} 
We evaluate the performance of EmbRace among four NLP models listed in Table \ref{tab:models}. We trained LM model with LM1B \cite{lm1b} dataset, GNMT-8 with  WMT-16 En-De \cite{wmt16}, Transformer with WMT-14 En-De \cite{wmt14} and BERT-base for question answering task with Squad \cite{squad}. The batch size per worker of LM, GNMT-8 and BERT-base are 128, 128 and 32 on RTX3090 GPUs, 128, 32 and 4 on RTX2080 GPUs, respectively. Since the input batch length is flexible in Transformer, we use max tokens per batch of 5120 on RTX3090 GPUs and 500 on RTX2080 GPUs. When measuring the training speed, we use tokens/sec as the metric, where we accumulate the non-padding words in each batch as the number of tokens. All reported throughputs are averaged over the same 500 training steps.

\subsubsection{Baselines} 
Altogether four approaches are compared with EmbRace in our experiments: 
(i) \textbf{BytePS} \cite{byteps}: a PS based distributed framework that integrates ByteScheduler \cite{bytescheduler} for partitioning tensors and scheduling communication with a priority queue, but it treats sparse tensors as dense tenors;
(ii) \textbf{Horovod AllReduce}: the popular distributed communication framework. In the PyTorch implementation of Horovod 0.21.3, the default communication method for sparse tensors is AllReduce;
(iii) \textbf{Horovod AllGather}: newly introduced with the Horovod 0.22.0 and becomes the default method for sparse data in its PyTorch part, where using AllGather to aggregate sparse tensors and AllReduce for dense tensors;
(iv) \textbf{Parallax}~\cite{parallax}: a hybrid communication approach that uses a partitioned PS for the sparse communications and AllReduce for the dense communications.

\subsection{End-to-End Training Performance}\label{sec:performance}
Figure \ref{fig:performance} shows the overall training throughputs for LM, GNMT-8, Transformer and BERT-base model on 4, 8 and 16 GPUs. 
EmbRace achieves a 1.02$\times$-1.77$\times$ speedup on RTX3090 nodes and 1.09$\times$-2.41$\times$ on RTX2080 nodes across four benchmark models. We present the experiment details among different models as follows. 

\textbf{LM}: Over the fastest baseline, EmbRace performs 1.18$\times$-1.77$\times$ better on RTX3090 GPUs and 1.99$\times$-2.41$\times$ on RTX2080 GPUs.
In our experiments, the LM model has the largest sparse parameter ratio and holds two large embedding tables, each taking over 1.5GB. So that dense communication methods (Horovod AllReduce and BytePs) are too slow. 
Also limited by the huge embedding tables and GPU memory, RTX3090 GPUs could hold the entire LM model but for RTX2080 GPU we have to put embedding tables on the CPU. 
For RTX3090 GPUs, the second-fastest method is Horovod AllGather due to its effective NCCL implementation.
For RTX2080 GPUs, Horovod AllGather is the second fastest method on 4 GPUs but is surpassed by Parallax on 8 and 16 GPUs, because the scalability of PS is better than AllGather, which is discussed in Section \ref{sec:spasrcomm}.

\textbf{GNMT-8/Transformer:}
EmbRace outperforms the best baseline 10.3\%-26.8\% in GNMT-8 and 11.5\%-17.6\% in Transformers with RTX3090 GPUs. With RTX2080 GPUs, EmbRace sees 9.4\%-29.6\% speedups in GNMT-8 and 10.5\%-28.2\% speedups in Transformers.
There are fewer sparse parameters than LM in translation models GNMT-8 and Transformer.
When training GNMT-8 and Transformer with RTX3090 GPUs, Horovod AllGather is the fastest method on 4 and 8 GPUs, but performs worse than Horovod AllReduce on 16 GPUs due to its scalability.
BytePs performs worse than Horovod AllReduce because BytePs uses share memory to speed up communication. In our hardware environment, the speed of RAMs is slow and would damage the performance of BytePs.
The Parallax is less effective than Horovod AllReduce due to the frequent memory copy between GPU and CPU.
However on RTX2080 GPUs, dense methods get poor performance due to the smaller batch size and lower intra-node bandwidth. Horovod AllGather is the fastest baseline on 4 and 8 GPUs while exceeded by Parallax on 16 GPUs.

\textbf{BERT-base:}
When compared to the best baseline,
EmbRace achieves 1.02$\times$-1.06$\times$ throughput on RTX3090 GPUs and 1.10$\times$-1.40$\times$ speedups on RTX2080 GPUs.
BERT-base holds the least number of parameters and the minimum ratio of computation to communication across benchmark models. In RTX3090 clusters, the BP process is long enough for Horovod AllReduce and BytePs to transmit a dense formatted embedding table. Hence Horovod AllReduce and BytePs become faster and 2D Communication Scheduling provides less performance improvement.  
As for RTX2080 clusters, the communication becomes an obvious bottleneck with the decreasing batch size thus EmbRace gains more accelerations. 

\begin{figure}[tb]
\centering
\subfigure[16 RTX3090 GPUs, 1.45$\times$-2.56$\times$]{
\includegraphics[width=0.48\linewidth]{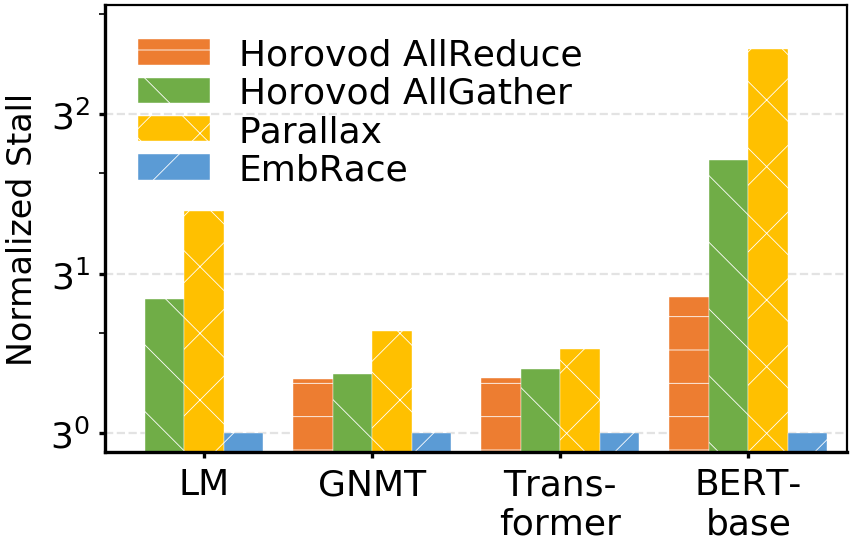}
}
\subfigure[16 RTX2080 GPUs, 1.37$\times$-3.02$\times$]{
\includegraphics[width=0.48\linewidth]{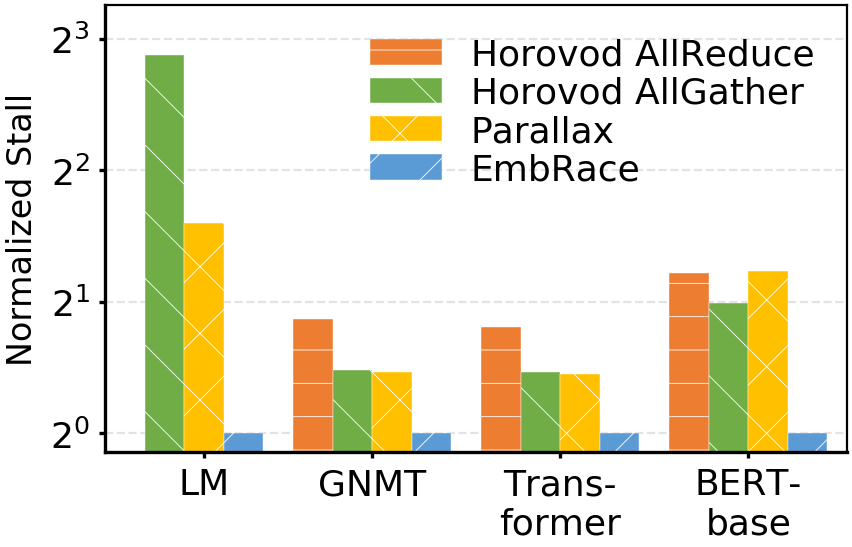}}
\caption {Computation Stall comparing of LM, GNMT-8, Transformer and BERT. Stall values are normalized by EmbRace.}
\label{fig:latency}
\end{figure}

\subsection{Communication Efficiency}

As EmbRace focuses on accelerating sparse communication in model training, the end-to-end training speed results might not be enough. We bring out another index, \textit{Computation Stall}, to further evaluate communication efficiency. In this paper, the Computation Stall is defined as the computation stall time caused by communication during the training procedure. For EmbRace, the Computation Stall consists of the Vertical Sparse Scheduling computation and communications that are not overlapped by computation. While for other approaches, Computation Stall is only the non-overlapping communication overheads. 

Computation Stall is mainly affected by three conditions: the network bandwidth, communication efficiency, and overlap ratio. Predictably, higher bandwidth and communication efficiency lead to a lower Computation Stall. With more unsatisfactory GPU performance or larger batch size, the computation time becomes longer and thus overlap ratio becomes higher, reducing the Computation Stall as well. In EmbRace, Sparsity-aware Hybrid Communication concentrates on boosting each sparse communication operation and 2D Communication Scheduling aims to overlap more communication with computation. We expect that these two techniques work together and reduce the Computation Stall greatly.

Figure \ref{fig:latency} shows the Computation Stall comparison, whose values are normalized by EmbRace, among four benchmark models on 16 GPUs. In LM, Horovod AllReduce gets such a large Computation Stall that it is hard to illustrate, so we omit it. 
EmbRace sees 1.45$\times$-2.56$\times$ speedups on 16 RTX3090 GPUs and 1.37$\times$-3.02$\times$ on 16 RTX2080 GPUs.
EmbRace performs surprisingly in LM and BERT-base, providing at least 49.7\%  reduction of Computation Stall. For GNMT-8 and Transformer, although the large amount of dense communication limits further accelerations, EmbRace does a good job in sparse communication and scheduling, cutting down a minimum of 26.4\% Computation Stall. 
The reduced Computation stalls lead to the end-to-end speedup of EmbRace. For example, when training BERT-base on RTX3090 GPUs, the batch size is larger, and the BP time can cover most communications. So computation stalls account for a low proportion of the overall time. EmbRace obviously reduces computation stalls, but the speedups in end-to-end performance are minor. While on RTX2080 GPUs, the computation stall becomes a major factor affecting training performance. The stall reductions of EmbRace bring more speedups to end-to-end performance.

\begin{figure}[tb]
\centering
\subfigure[16 RTX3090 GPUs]{
\includegraphics[width=0.48\linewidth]{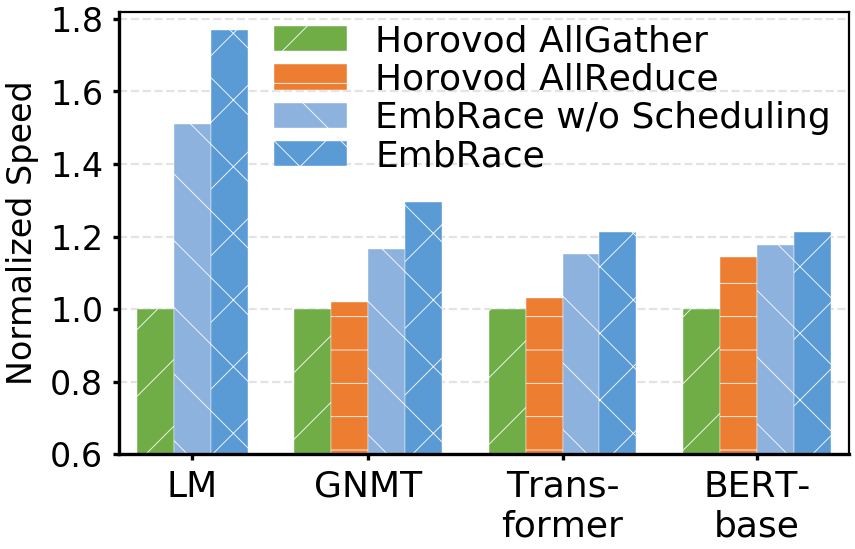}
}
\subfigure[4 RTX3090 GPUs]{
\includegraphics[width=0.48\linewidth]{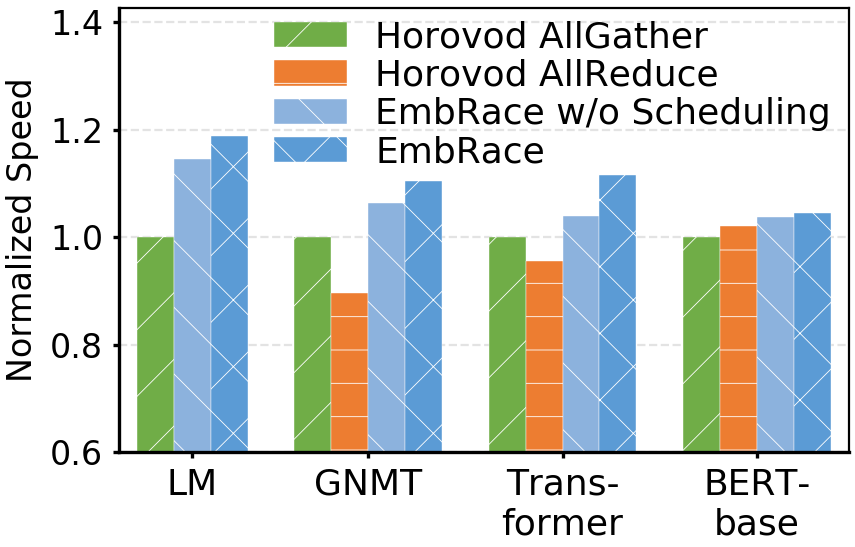}}
\caption {Ablation experiments on the optimizations of EmbRace with 16 and 4 RTX3090 GPUs. Training speeds are normalized by Horovod AllGather.}
\label{fig:ablation}
\end{figure}

\subsection{Ablation Study}\label{sec:ablation}
To analyze the effectiveness of our optimizations, we carry out the ablation study with all four benchmark models on RTX3090 GPUs. When compared with Horovod AllGather and Horovod AllReduce, 
results of EmbRace without Scheduling could show the performance gains of Sparsity-aware Hybrid Communication. To demonstrate the effect of 2D Communication Scheduling, we could compare the training speeds of EmbRace with and without Scheduling.

Figure \ref{fig:ablation} presents the experiment results, where training speed values are normalized by Horovod AllGather. Sparsity-aware Hybrid Communication provides 2.9\%-51.0\% speedups and 2D Communication Scheduling brings another 3.0\%-26.0\% speedups on 16 GPUs. 
On 4 GPUs, Sparsity-aware Hybrid Communication offers 1.5\%-14.6\% speedups and 2D Communication Scheduling sees another 0.7\%-7.5\% speedups. 
Both techniques of EmbRace could bring remarkable improvements. With the increasing number of GPUs, communication accelerations become more obvious.

\begin{figure}[tb]
\centering
\subfigure[LM]{
\includegraphics[width=0.475\linewidth]{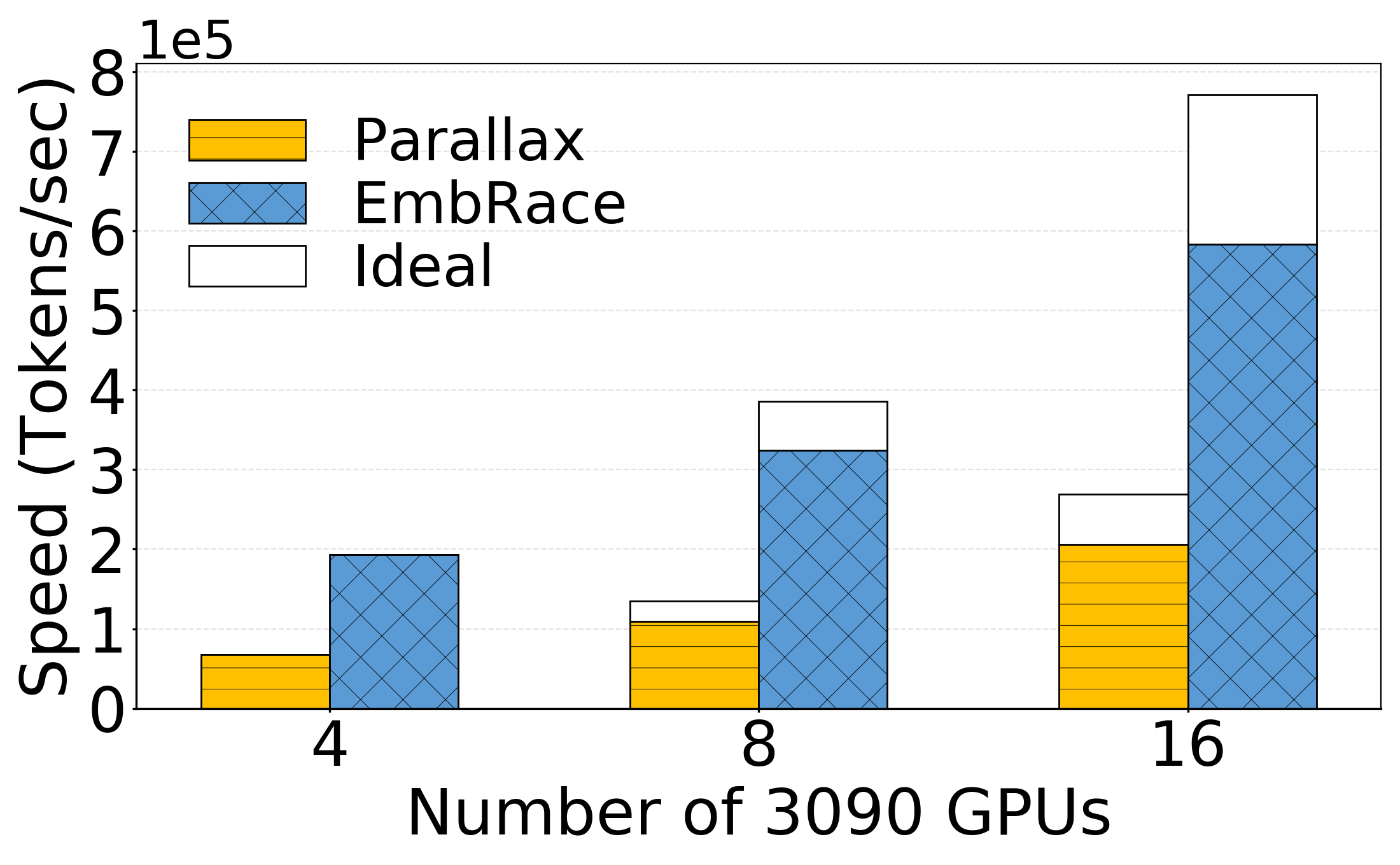}
}
\subfigure[GNMT-8]{
\includegraphics[width=0.475\linewidth]{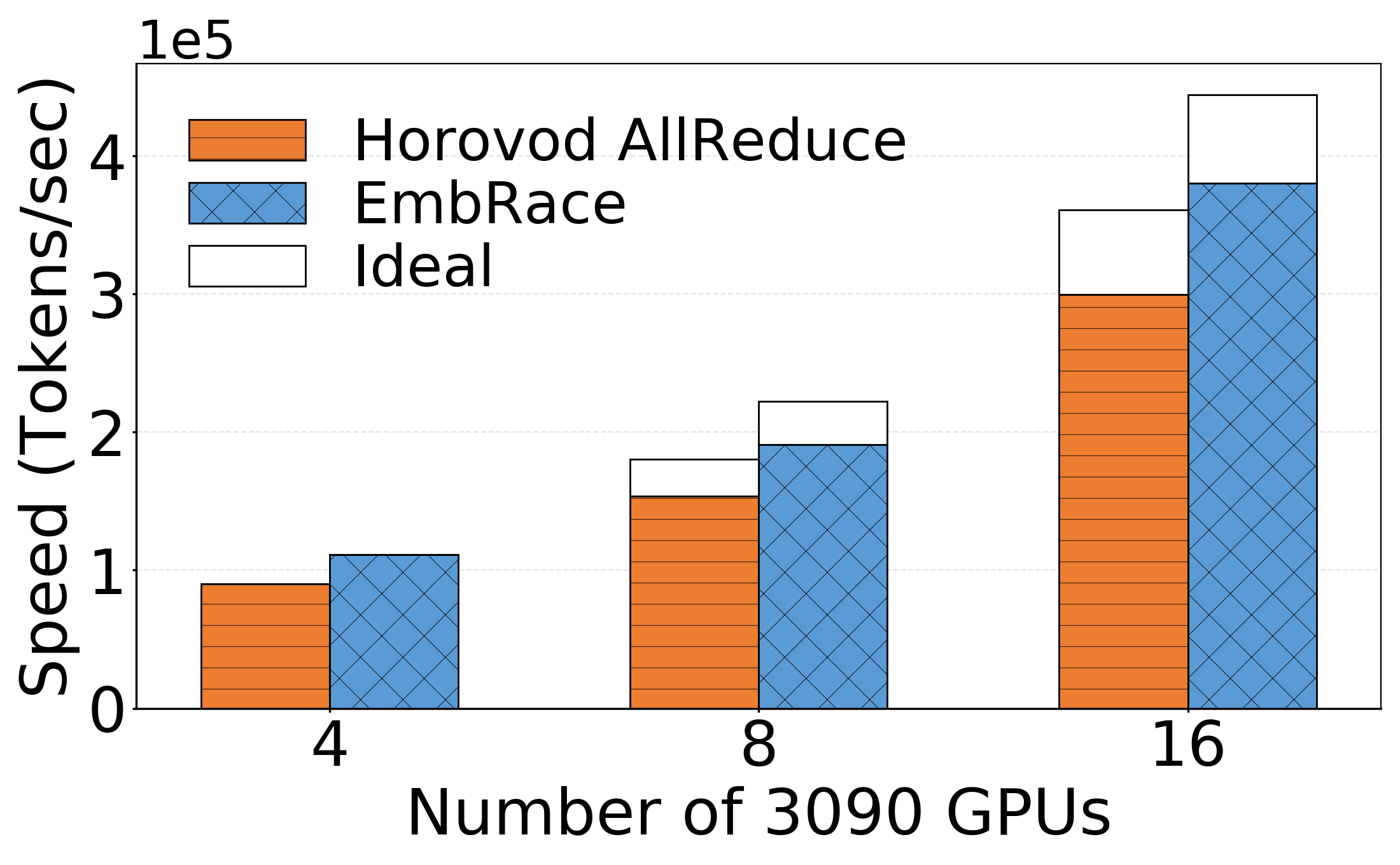}
}
\subfigure[Transformer]{
\includegraphics[width=0.475\linewidth]{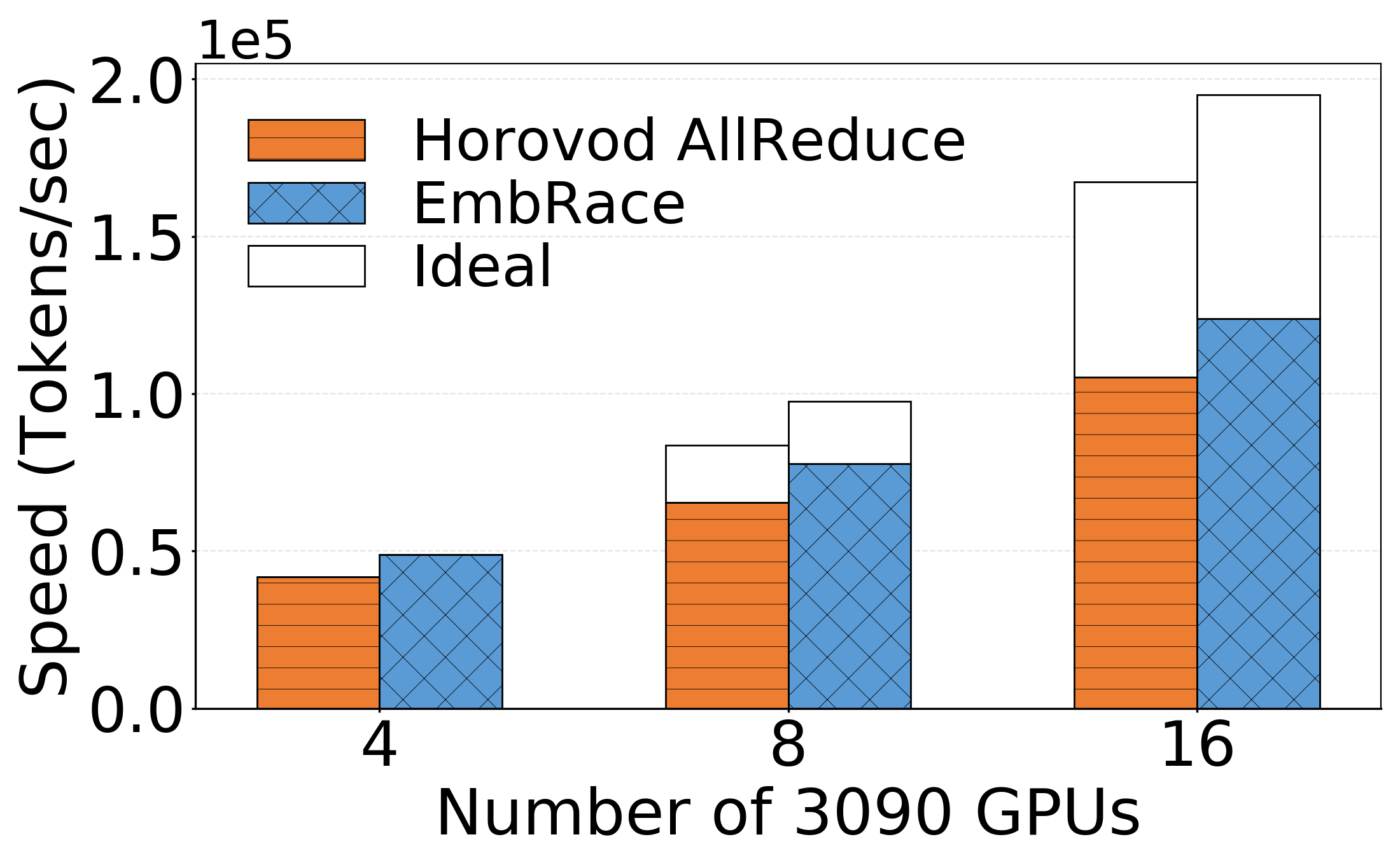}
}
\subfigure[BERT-base]{
\includegraphics[width=0.475\linewidth]{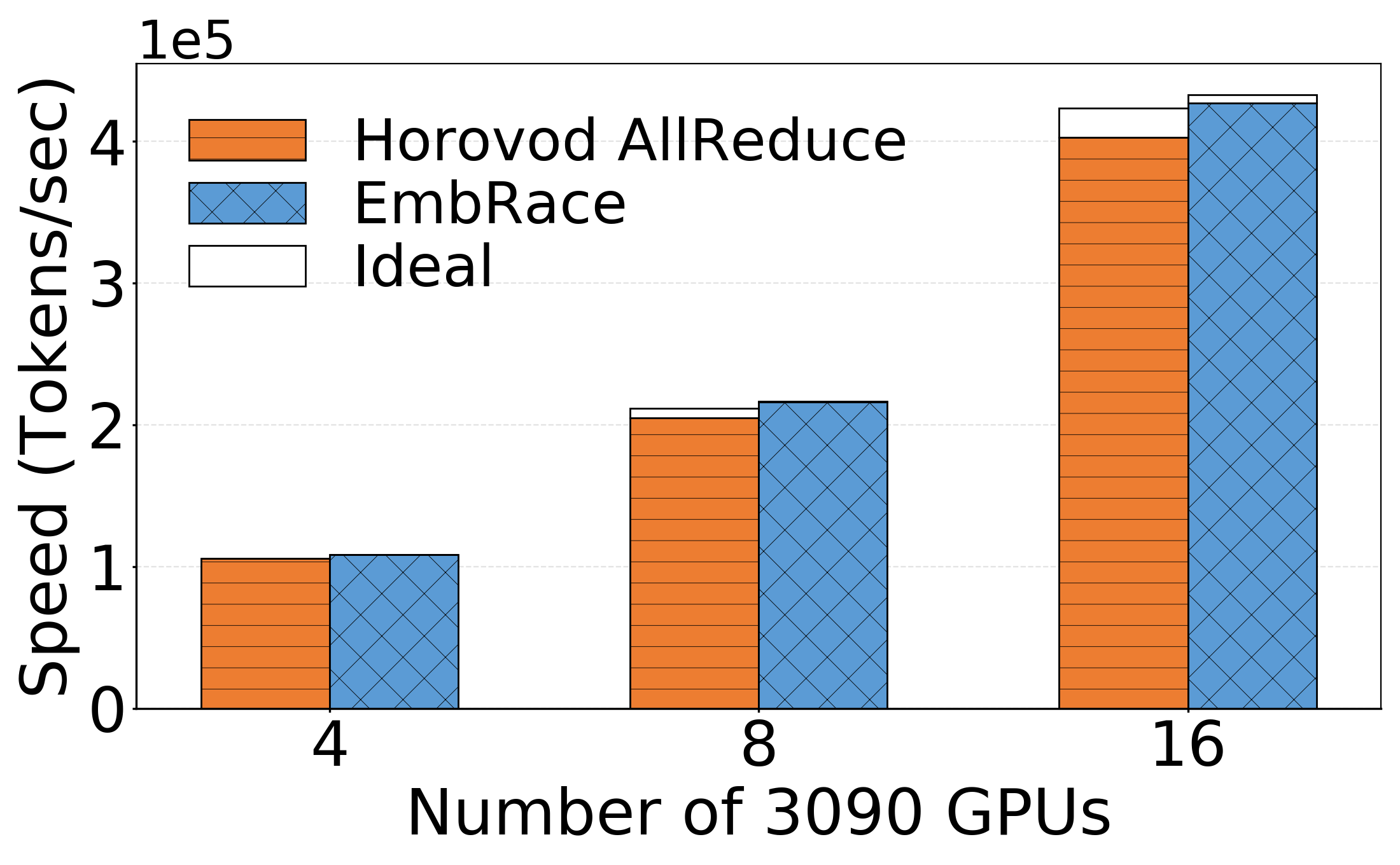}
}
\caption{Scaling performance on RTX3090 GPUs, compared to the approach with the second-best scalability.}
\label{fig:scalability}
\end{figure}

\subsection{Scalability}

To demonstrate the scalability of EmbRace, we compare it with Horovod AllReduce which owns the best scalability in GNMT-8, Transformer and BERT-base.
For the LM model, we choose Parallax as the competitor since dense methods are extremely slow, and PS architecture maintains good scalability in our analysis.

Figure \ref{fig:scalability} shows the training throughputs compared with corresponding ideal linear scaling results. We use the related throughputs of models on 4 RTX3090 GPUs as the benchmark, then calculate the ideal linear results of 8 and 16 GPUs based on the benchmark.
When scaled from 4 GPUs to 16 GPUs across GNMT-8, Transformer and BERT-base, EmbRace offers 3.42$\times$, 2.53$\times$ and 3.94$\times$ speedups while Horovod AllReduce gets 3.32$\times$, 2.51$\times$ and 3.81$\times$, respectively. 
When scaling LM to 16 GPUs, EmbRace gets 3.14$\times$ throughput, where Parallax gets 3.06$\times$.
Due to the limitation of the number of devices, we did not test on more server nodes. With the help of better scalability, we expect that EmbRace will have more significant advantages on more GPUs.

\begin{figure}[tb]
\centering
\subfigure[PPL vs. Steps in LM]{
\includegraphics[width=0.475\linewidth]{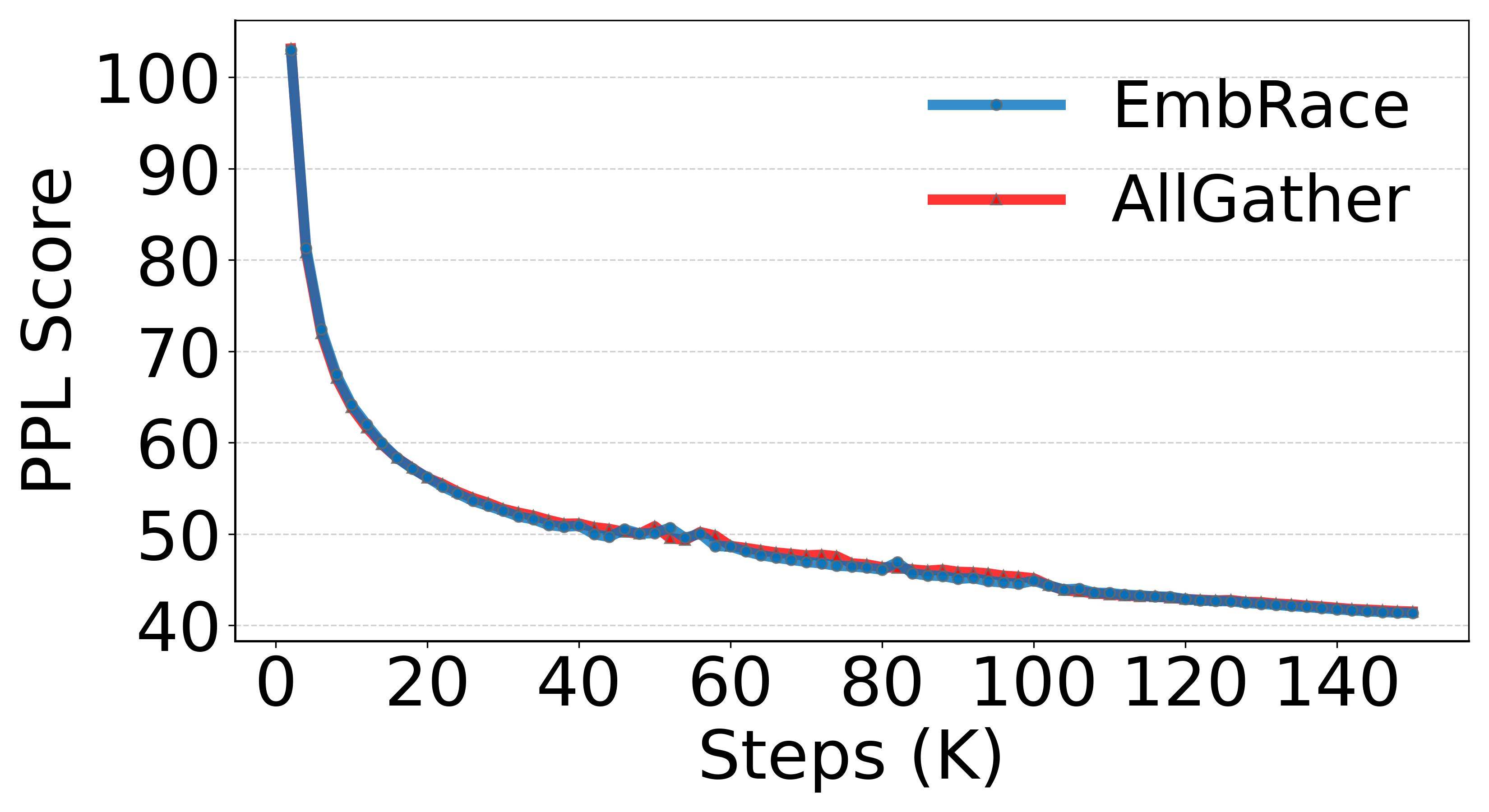}
}
\subfigure[BLEU vs. Epochs in GNMT-8]{
\includegraphics[width=0.475\linewidth]{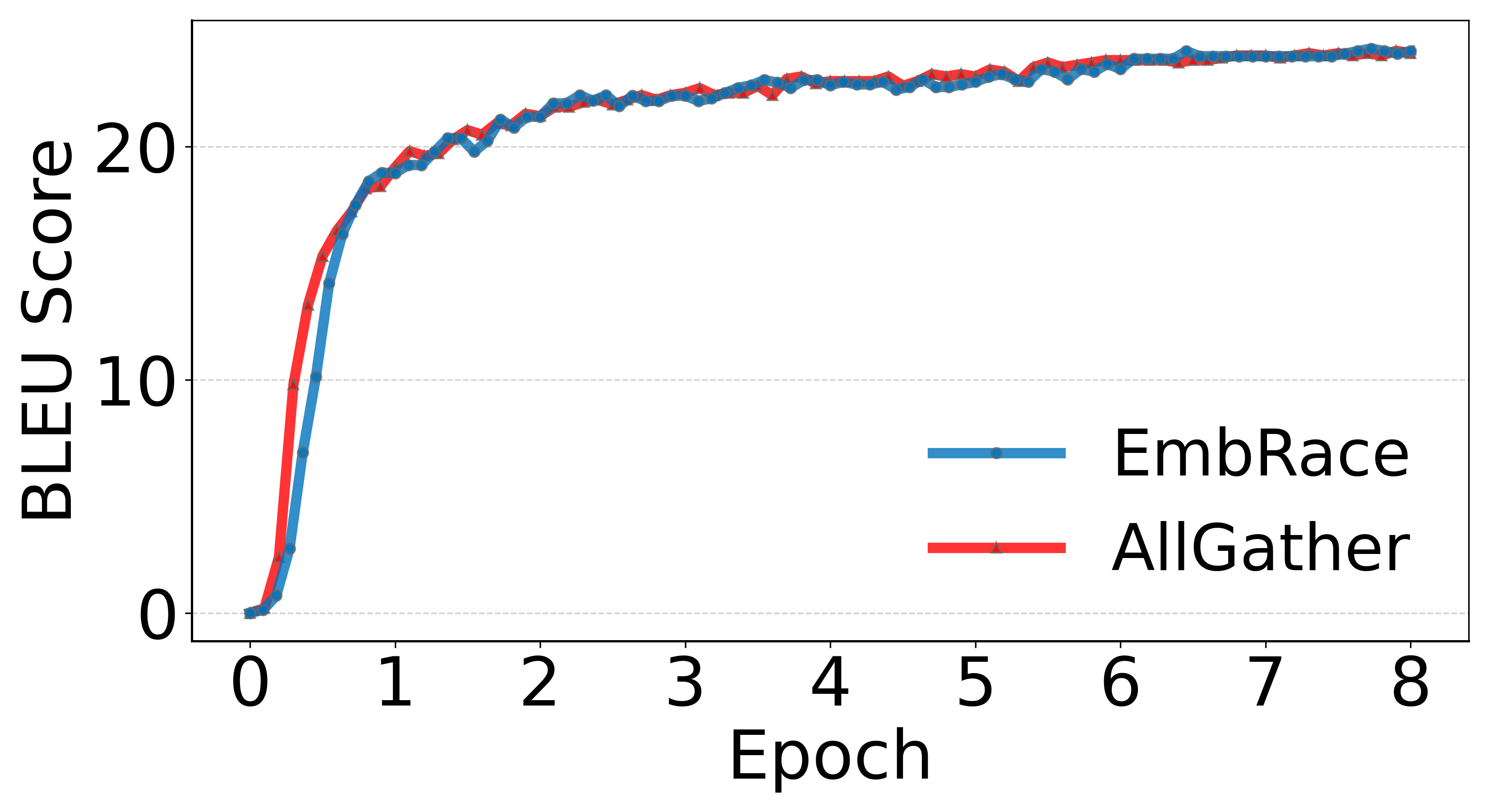}
}

\caption{Convergence comparing between EmbRace and Horovod AllGather on 8 RTX3090 GPUs.}
\label{fig:convergency}
\end{figure}

\subsection{Convergence}
Since EmbRace is a synchronous data-parallel training approach, it should correctly converge models like other synchronous approaches. However with Vertical Sparse Scheduling, each sparse gradient is divided into two parts and requires parameter updating twice with sparse optimizers. 
Because the common sparse optimizer such as Adagrad \cite{adagrad} and SGD \cite{sgd} is fully element-wise, no matter whether updating embedding matrices with multiple gradient parts or a whole gradient would lead to the same result.
But we used Adam \cite{cite20} optimizer in all experiments. 
Most parts of Adam are element-wise except the state parameter \textit{step}, which would be accumulated after every optimizing call, leading to a minor difference between a single or two parameter updates. 
Therefore, we modify the Adam optimizer in PyTorch, updating the \textit{step} state only at applying the delayed sparse gradients to embedding parameters. This modification ensures synchronous training and the rate of convergence. 

Figure \ref{fig:convergency} illustrates the convergence curves of EmbRace and Horovod AllGather in LM and GNMT-8 with 8 RTX3090 GPUs. We trace the perplexity (PPL) scores for LM with training steps and BLEU scores for GNMT-8 along with epoch numbers. As shown in Figure \ref{fig:convergency}(a) and Figure \ref{fig:convergency}(b), both methods converge the model into PPL 41.5 and BLEU 24.0 in similar numbers of training iterations or epochs.  

\section{Related Work}

\textbf{Communication acceleration.}
Upon popular synchronous data-parallel frameworks such as Horovod \cite{cite14} and PyTorch Distributed \cite{cite13}, there are different directions for accelerating communications: (1) optimizing multiple GPU intra-machine communication \cite{cite29}; (2) applying topology-aware hierarchical collective communication \cite{cite22,cite31}; (3) reducing messages size with gradient compression \cite{cite7,cite32}; (4) speeding up individual messages with specific networks such as RDMA \cite{cite28}.
These works concentrate on each communication operation and could be orthogonal and complementary to EmbRace.

\textbf{Sparse communication in DNN training.}
To improve sparse communication performance in DNN training, Horovod and PyTorch Distributed adopt AllGather for sparse communications; Parallax \cite{parallax} uses architecture combining PS with AllReduce; HMA \cite{hma} integrates model average with AllReduce; OmniReduce \cite{omni} implements a sparsity-aware AllReduce algorithm; and S2 reducer \cite{s2reducer} designs a compress algorithm that adapts sparse tensors to AllReduce primitives. However, they do not utilize the property of embedding matrix, which could reduce the communication overhead further. 

\textbf{Communication scheduling.}
On top of the wait-free backward propagation \cite{poseidon} which is supported by most DL frameworks, communication scheduling could further overlap communication with FP computation.
TicTac \cite{tic} attempts to schedule communication in PS architecture with a priority queue;
P3 \cite{p3} proposes a similar idea and implements layer partitioning for scheduling in a layer-wise granularity;
ByteScheduler \cite{bytescheduler} expands communication scheduling to AllReduce architecture and adopts parameter partitioning for a finer granularity;
PACE \cite{pace} changes the priority queue into a preemptive queue in AllReduce and uses tensor fusion for better bandwidth usage.
However, these methods only work for dense data. In EmbRace, we benefit from these approaches but we take sparsity and characteristics of NLP models into consideration.

\section{Conclusion}
In this paper, we present EmbRace, an efficient distributed sparse communication framework for NLP model training.
EmbRace introduces Sparsity-aware Hybrid Communication that uses AlltoAll primitive with model parallelism to race the communication of embedding tables. We also design 2D Communication Scheduling which embraces the Horizontal and Vertical Scheduling, letting EmbRace optimizes model computation procedure, utilizes GPU idle time, and achieves a thorough overlapped communication schedule.
Experiments show EmbRace achieves up to 66.7\% Computation Stall reduction, up to 2.41$\times$ training speedup, and better scalability, when compared to the best baseline.
Although training giant NLP models with thousands of GPUs becomes a trend nowadays, training models swiftly with limited resources such as RTX2080 GPUs still matter. And EmbRace could benefit sparse communications in giant NLP models training as well.

\begin{acks}
This work is sponsored by National Key R\&D Program of China (2021YFB0301200). Zhiquan Lai is the corresponding author of this paper.
% This work is sponsored in part by the National Key Research and Development Program of China under Grant No.2018YFB0204300, and the National Natural Science Foundation of China under Grant No.62025208. 

\end{acks}

%%
%% The next two lines define the bibliography style to be used, and
%% the bibliography file.
\bibliographystyle{ACM-Reference-Format}
\bibliography{sample-base}

\end{document}